\documentclass[10pt]{article}
\usepackage[english]{babel}

\usepackage{amsthm}
\usepackage{amsmath}
\usepackage[hidelinks]{hyperref} 
\usepackage{graphicx} 
\usepackage[utf8]{inputenc}
\usepackage[dvipsnames,x11names]{xcolor}

\usepackage{makecell}
\usepackage{tabularx}
\usepackage{subfig}

\usepackage{bm}
\usepackage{multirow}
\usepackage{amssymb}
\usepackage{float}
\usepackage{units}

\usepackage[b5paper, twoside]{geometry}
\usepackage{titlesec}

\usepackage[longnamesfirst, round]{natbib}
\setcitestyle{authoryear,open={(},close={)}}

\numberwithin{equation}{section}
\theoremstyle{plain}

\theoremstyle{definition}

\usepackage{authblk}

\newlength{\currentparskip}
\newlength{\currentparindent}
\newenvironment{myabstract}[1]
    {
    \setlength{\currentparskip}{\parskip}
    \setlength{\currentparindent}{\parindent}
    \begin{center}
    \begin{minipage}[]{#1}
    \setlength{\parskip}{\currentparskip}
    \setlength{\parindent}{\currentparindent}
    \small
    \begin{center} \textbf{Abstract} \end{center} \vspace*{-0.5em} \vspace*{-\currentparskip}
    }
    {
    \end{minipage}
    \end{center}
    }
    
\newenvironment{mykeywords}[1]
    {
    \vspace*{-1.5em}
    \begin{center}
    \begin{minipage}[]{#1}
    \small
    \noindent \textbf{Keywords:}
    }
    {
    \end{minipage}
    \end{center}
    }

\begin{document}

\graphicspath{{figures/}}

\title{Learning Physics between Digital Twins with Low-Fidelity Models and Physics-Informed Gaussian Processes}

\author[1]{Michail Spitieris}
\author[1]{Ingelin Seinsland}
\affil[1]{{Department of Mathematical Sciences, NTNU, Norway}}
\date{}

\maketitle

\begin{myabstract}{0.85\linewidth}

\noindent
A digital twin is a computer model that represents an individual, for example, a component, a patient or a process. In many situations, we want to gain knowledge about an individual from its data while incorporating imperfect physical knowledge and also learn from data from other individuals. 
In this paper, we introduce a fully Bayesian methodology for learning between digital twins in a setting where the physical parameters of each individual are of interest. 
A model discrepancy term is incorporated in the model formulation of each personalized model to account for the missing physics of the low-fidelity model.
To allow sharing of information between individuals, we introduce a Bayesian Hierarchical modelling framework where the individual models are connected through a new level in the hierarchy. 
Our methodology is demonstrated in two case studies, a toy example previously used in the literature extended to more individuals and a cardiovascular model relevant for the treatment of Hypertension. 
The case studies show that 1) models not accounting for imperfect physical models are biased and over-confident, 2) the models accounting for imperfect physical models are more uncertain but cover the truth, 3) the models learning between digital twins have less uncertainty than the corresponding independent individual models, but are not over-confident.
\end{myabstract}
\vspace{0.2cm}

\begin{mykeywords}{0.85\linewidth}
Digital twin, Gaussian process, physics-informed ML, Bayesian Hierarchical models, Bayesian calibration, model discrepancy, inverse problem
\end{mykeywords}

\section{Introduction}

A digital twin can be defined as a virtual representation of a physical asset enabled through data and simulators \citep{DigitalTwinAdilogTrond}. Simulators refer to mathematical models, and physical models based on first principles are often preferred to incorporate system knowledge and gain explainability.  A known challenge of deployment of digital twins is scalability,  i.e. the ability to  accessible robust digital twin implementations at scale \citep{DigitalTwinAdilogTrond, DigitalTwinScale}. The ability to make better decisions is the motivation for digital twins.
In this paper, we consider the case that better decisions can be made with knowledge about the physical parameters of an imperfect
physical model. 
To infer the physical parameters, noisy observed data are typically used. 
Another source of uncertainty in applications of low-fidelity physical models is the model-form uncertainty that arises from imperfect physical models.

\begin{figure}[t!]
    \centering
    \includegraphics[scale=0.55]{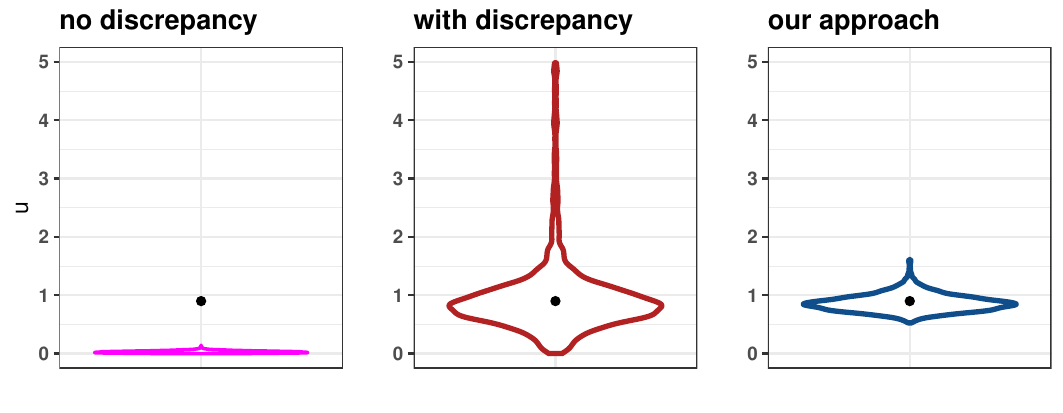} 
    \vspace{-0.5cm}
	\caption{Posterior distributions of the physical parameter u for three modelling approaches when there is systematic discrepancy between the modelling choice and the reality. The posterior distribution of the model that does not account for model discrepancy (left), the model that accounts for discrepancy but does not share information between individuals (middle) and the proposed approach (right).}  \label{fig:intro}
\end{figure}

The primary motivation of this work is a medical Digital Twin aiming at preventing and treating hypertension (or high blood pressure). Physical models of the cardiovascular system are based on physical parameters that can not be measured directly and are important to hypertension. However, these parameters can be inferred by fitting the physical model to the observed data. 
Furthermore, the models can be used to predict the development of hypertension under different treatments.
In one of our case studies, a low-fidelity model for the cardiovascular system, the Windkessel model \citep{westerhof2009arterial} is used. This model is a differential equation linking blood flow and blood pressure and has two physically interpretable parameters, the arterial compliance $C$ and resistance $R.$ The estimated values of the parameters can direct the treatment of hypertension for patients. The parameters vary between patients, and we aim to estimate these using noisy flow and pressure observations. It is known that low-fidelity models produce biased parameter estimates if we don't account for model discrepancy \citep{brynjarsdottir2014learning}, as can also be seen in Figure \ref{fig:intro}, left. 

Recently, \citet{spitieris2023bayesian} introduced and demonstrated a methodology for Bayesian analysis of imperfect physical models using physics-informed Gaussian process (GP) priors. 
They combine a model formulation that accounts for discrepancy introduced by \citet{kennedy2001bayesian} with Physics-informed GP priors for physical models described by differential equations \citep{raissi2017machine}.
This gives computational efficiency and uncertainty quantification of the physical parameters of interest through a fully Bayesian approach.

For each digital twin accounting for model discrepancy combined with noisy data can result in too uncertain estimates to be of practical use (see Figure \ref{fig:intro}, middle). Further, there can be identification issues related to the discrepancy. The working hypothesis is that the uncertainty of the parameters for the individual digital twins can be reduced by using information from other digital twins 
(see Figure \ref{fig:intro}, right).

In this paper, we introduce a framework for how individual digital twins can share data to reduce uncertainty. The physical models are cast into a Bayesian hierarchical model framework which is extended with an extra level to allow both the physical parameters and the discrepancy to gain information from the other individuals. 
For learning the imperfect physics between individuals, two approaches are introduced, one that  assumes the same GP parameters for the discrepancy for all individuals and one that can be seen as using a prior learned from all individuals for the parameters of the GPs
representing the discrepancies. 

The remainder of the paper is organized as follows. In Section \ref{sec:background}, we review the main components of our proposed methodology, that is the physics-informed priors and the Bayesian calibration accounting for model discrepancy. In Section \ref{sec:method},  we introduce the proposed methodology. In Section \ref{sec:toy}, we consider a synthetic case study with a toy example that has been previously used in the literature but now is extended for more individuals. In Section \ref{sec:WK_synth}, we consider a synthetic case study with a cardiovascular model where we simulate data for a more complex model than our modelling choice, and then we also fit the model to real data obtained from a pilot randomized controlled trial study. Finally, we discuss the results, limitations and broader impact of our work. 
The code to replicate all the results in the paper is available at \url{https://github.com/MiSpitieris/Learning-Physics-between-DTs}. 
\begin{figure*}[h!]
    \centering
    \includegraphics[scale=0.5]{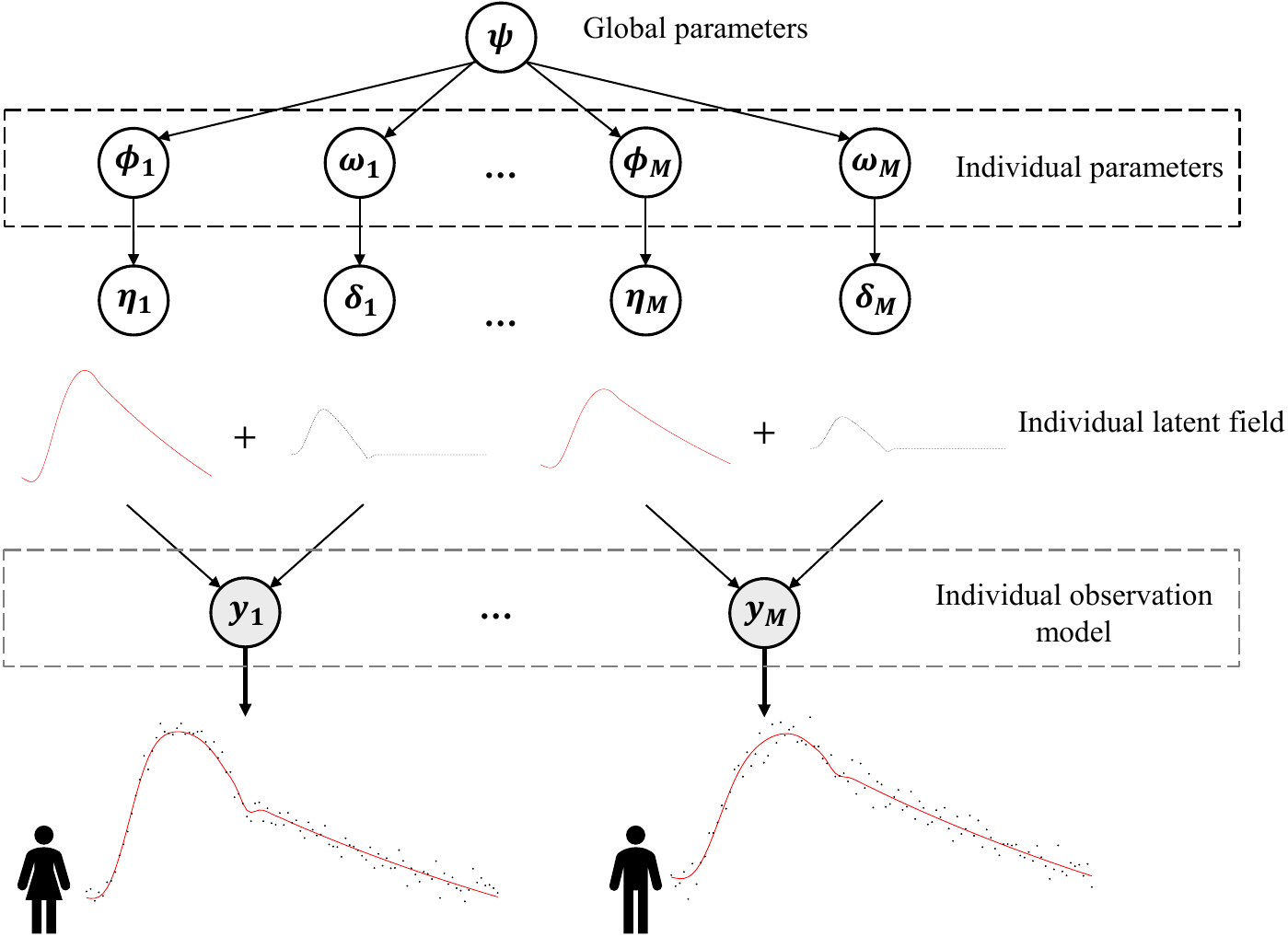} 
    \caption{The hierarchical model detailed in Section \ref{sec:method} represented as directed acyclic graph (DAG). It presents the general case where the personalized physical model for the first and M-th individual is denoted as $\boldsymbol{\eta}_1 = \eta(\mathbf{X}_1, \boldsymbol{\phi}_1)$ and $\boldsymbol{\eta}_M = \eta(\mathbf{X}_M, \boldsymbol{\phi}_M)$ respectively and we are interested in learning the individual physical parameters $\boldsymbol{\phi}_1, \ldots, \boldsymbol{\phi}_M.$ In the individual latent field row, the red line represents the posterior mean of the (low fidelity) latent physical model and the black dashed line is the posterior mean of the latent discrepancy model. At the bottom, we have the posterior mean of the bias-corrected (low fidelity) model, where the black dots are the observed data.}  \label{fig:DAG}
\end{figure*}

\section{Background} \label{sec:background}

\subsection{Accounting for Model Discrepancy} \label{sec:MD}

Let $\mathbf{x}=(x_{1},\ldots, x_{k})$ be the $k$ input variables, and $\boldsymbol{\phi}=(\phi_1,\ldots,\phi_p)$ the vector of unknown parameters for the physical model $\eta(\bf{x}, \boldsymbol{\phi})$. This model is typically a simplified representation of the reality $\mathcal{R},$ and does not fit the observed data well, resulting in biased physical parameters $\boldsymbol{\phi}$ estimates \citep{brynjarsdottir2014learning}. To account for this model form uncertainty, \citet{kennedy2001bayesian} (KOH) suggested a Bayesian calibration framework which incorporates a functional model discrepancy  in the model formulation. More specifically, the noise corrupted observed data $\mathbf{y}^\mathcal{R}$ are described by the physical model $\eta$ and the systematic model discrepancy $\delta$ as follows
\begin{equation}\label{eq:KOH}
   y^R(\mathbf{x}) = \eta(\mathbf{x},\boldsymbol{\phi})+\delta(\mathbf{x})+\varepsilon,
\end{equation}
where $\varepsilon$ is the noise term. Since the discrepancy is an unknown function of the inputs $\mathbf{x},$ KOH used a flexible Gaussian process prior \citep{rasmussen2003gaussian} to model the discrepancy, $\delta(\mathbf{x})\sim GP(0,K_{\delta}(\mathbf{x}, \mathbf{x}'\mid \boldsymbol{\omega})),$ where $K_\delta$ is the covariance function and $\boldsymbol{\omega}$ are kernel hyperparameters. The physical model is usually computationally expensive, and thus KOH replaced the physical model numerical simulator with an emulator, which is a GP model trained on $N$ simulator output runs trained on an experimental design on the input and parameter space. This results in a GP model which utilizes the two sources of information, $N$ simulator runs data and $n$ observed data and has a computational complexity of $\mathcal{O}((N+n)^3),$ where typically $N\gg n.$ If the deterministic model is fast to evaluate, the computational complexity reduces. If  an i.i.d. Gaussian error term, $\varepsilon \sim N(0,\sigma^2)$ is assumed we get the following model formulation \citep{higdon2004combining} 
\begin{equation}\label{eq:higdon}
    y^\mathcal{R} \sim GP(\eta(\mathbf{x}, \boldsymbol{\phi}), K_\delta(\mathbf{x},\mathbf{x'}\mid \boldsymbol{\omega})+\sigma^2),
\end{equation}
and this formulation will be used in Section \ref{sec:toy}. To infer model parameters ($\boldsymbol{\omega}, \boldsymbol{\phi}, \sigma$) in equation  \eqref{eq:higdon}, Bayesian inference is used by assigning priors to the model unknowns and specifically priors that reflect underlying knowledge about $\boldsymbol{\phi},$ and the posterior distribution is sampled using MCMC.

\subsection{Physics-Informed (PI) Priors} \label{sec:PIP}
We focus on the construction of physics-informed Gaussian process priors for physical models that are described by linear differential equations $\mathcal{L}_{x}^{\boldsymbol{\phi}}u(x) = f(x),$ where $\mathcal{L}$ is the linear differential operator and $\boldsymbol{\phi}$ the vector of physical parameters. To simplify the notation, we take $x$ to be univariate, however, the following results apply to higher dimensions,  e.g. see examples in \citep{raissi2017machine, spitieris2023bayesian}.

Suppose we have observations of the functions $u$ and $f$ at potentially different  locations $x_{u1},\ldots, x_{un_u}$ and $x_{f1},\ldots, x_{fn_f},$ where $n_u$ and $n_f$ are the number of data for the functions $u$ and $f$, respectively. The corresponding observed data are $\mathbf{y}_u = (y_{u1},\ldots,y_{un_u})$ and $\mathbf{y}_f = (y_{f1},\ldots,y_{fn_f}).$
The observed data are functional, and usually, $u(x)$ is a smooth function. 
Thus it is often reasonable to assume a GP prior, $u(x)\sim GP(0,K_{uu}(x, x'\mid \boldsymbol{\theta}))$ to describe the $u(x)$. 
The derivative of a GP is also a GP \citep{adler2010geometry} and, more specifically 
$\textrm{Cov}\left(u(x), \frac{\partial u(x')}{\partial x'}\right) = \frac{\partial K_{uu}(x,x'\mid \boldsymbol{\theta})}{\partial x'}$ and
$\textrm{Cov}\left(\frac{\partial u(x)}{\partial x}, \frac{\partial u(x')}{\partial x'}\right) = \frac{\partial^2 K_{uu}(x,x'\mid \boldsymbol{\theta})}{\partial x \partial x'},$ if the kernel is differentiable. Thus a convenient choice (but not the only) is the squared exponential kernel since it is infinitely differentiable.

\citet{raissi2017machine} utilized this result to build physics-informed priors for $\mathcal{L}_{x}^{\boldsymbol{\phi}}u(x) = f(x)$ by assuming a GP prior on $u(x).$ Then we have that $f(x) \sim GP(0, K_{ff}(x,x'\mid \boldsymbol{\theta},\boldsymbol{\phi})),$ where $K_{ff}(x,x'\mid \boldsymbol{\theta},\boldsymbol{\phi})=\mathcal{L}_{x}^{\boldsymbol\phi}\mathcal{L}_{x'}^{\boldsymbol\phi}K_{uu}(x,x'\mid \boldsymbol{\theta})$, and also
$K_{uf}(x,x'\mid \boldsymbol{\theta},\boldsymbol{\phi}) =\mathcal{L}_{x'}^{\boldsymbol\phi}K_{uu}(x,x'\mid \boldsymbol{\theta})$ and 
$K_{fu}(x,x'\mid \boldsymbol{\theta},\boldsymbol{\phi}) =\mathcal{L}_{x}^{\boldsymbol\phi}K_{uu}(x,x'\mid \boldsymbol{\theta}).$ This results in a multi-output (of $u$ and $f$) GP model that satisfies the differential equation. The physical parameters $\boldsymbol{\phi}$ have become hyperparameters of the physics-inspired covariance matrix, and therefore standard tools for inference can be used, where \citet{raissi2017machine} used maximum likelihood and obtained point estimates.

This approach was extended in a Bayesian framework \citep{spitieris2023bayesian}, which also accounts for model discrepancy in a model formulation similar to equation \eqref{eq:KOH}. More specifically, for noise corrupted data $y_u$ and $y_f$, we assume a discrepancy function on $u(x),$ and we have the following model formulation
\begin{equation*} 
  \begin{split}
    y_u & = u(x_u) + \delta_u(x_u) + \varepsilon_u, \\
	y_f &= f(x_f) + \varepsilon_f,
  \end{split}
\end{equation*}
where $\delta_u(x) \sim GP(0, K_{\delta}(x,x'\mid \boldsymbol{\omega}))$ and $\varepsilon_u\sim N(0,\sigma_u^2),$ $\varepsilon_f\sim N(0,\sigma_f^2)$ are the error terms. By assuming that $u(x)\sim GP(\mu_u(x_u \mid \boldsymbol \beta),K_{uu}(x, x'\mid \theta)),$ where $\mu_u$ is a mean function with parameters $\boldsymbol \beta,$  and consequently $\mu_f(x_f \mid \boldsymbol \beta, \boldsymbol \phi) = \mathcal{L}_{x}^{\boldsymbol{\phi}}\mu_u(x_u \mid \boldsymbol \beta)$, we get the following multi-output GP that accounts for model discrepancy 
\begin{equation} 
	p(\mathbf{y}\mid \boldsymbol{\beta}, \boldsymbol{\theta}, \boldsymbol{\omega}, \boldsymbol \phi, \sigma_u, \sigma_f) = \mathcal{N}(\boldsymbol{\mu}, \mathbf{K}_{\text{disc}}+\mathbf{S}) \label{eq:PI}
\end{equation}
where
$
\bf{y} = \begin{bmatrix}  \bf{y}_u \\  \bf{y}_f \end{bmatrix}
$, 
$
\mathbf{K}_\text{disc} =
\begin{bmatrix}
K_{uu} +K_\delta  & K_{uf}\\
K_{fu} & K_{ff}
\end{bmatrix}
$,
$
\mathbf{S} =
\begin{bmatrix}
\sigma_u^2 I_u & 0\\
0 & \sigma_f^2 I_f
\end{bmatrix} 
$ and 
$
\boldsymbol \mu =
\begin{bmatrix}
\boldsymbol \mu_u(\mathbf{X}_u \mid \boldsymbol \beta)\\
\boldsymbol \mu_f(\mathbf{X}_f \mid \boldsymbol{\beta,\phi})
\end{bmatrix}.
$ 
Finally, to infer model hyperparameters Hamiltonian Monte Carlo (HMC) \citep{neal2011mcmc} is used and more specifically, the NUTS variation \citep{hoffman2014no}. 

For the rest of the paper, we denote $\mathbf{X}= (\mathbf{X}_u, \mathbf{X}_f),$ where $\mathbf{X}$ can be multivariate and eq. \eqref{eq:PI} can be written as
$
	p(\mathbf{y}\mid \boldsymbol{\beta}, \boldsymbol{\theta}, \boldsymbol{\omega}, \boldsymbol \phi, \sigma_u, \sigma_f) = 
	\mathcal{N}(\boldsymbol{\mu} (\mathbf{X}\mid \boldsymbol{\beta,\phi}), 
	\mathbf{K}_{\text{disc}}(\mathbf{X}, \mathbf{X}\mid \boldsymbol \theta, \boldsymbol{\omega}, \boldsymbol \phi, \sigma_u, \sigma_f)+\mathbf{S}).
$

\section{Hierarchical Physical Models Accounting for Model Discrepancy}\label{sec:method}

In this section, the model and inference method for sharing information between individuals about parameters in physical models and discrepancy  are introduced. We assume that we have data from M individuals, 
where $m=\{1,\ldots,M\}$ denotes
the individual id, and $\mathbf{X}_m$ and  $\mathbf{y}_m$
the corresponding matrix of observed input data and  vector output for individual $m.$

Note that in the case of physics-informed priors, the vectors $\mathbf{X}_m$ and  $\mathbf{y}_m$ are joint for both the part related to $u$ and $f$ for individual $m$ as described in Section 2.2.
Our aim is to infer the physical parameters for each individual, denoted $\boldsymbol{\phi}_m.$

\subsection{Hierarchical Physical Model, Shared Global Parameters}
We now set up our most general hierarchical model where the individuals are connected through distributions of the conditional priors of the individual parameters. The model is illustrated in Figure \ref{fig:DAG} and mathematically formulated below: 
\begin{align} %
& \text{ \emph{Individual likelihood}: } &
\mathbf{y}_m & \mid \boldsymbol{\eta}_m, \boldsymbol{\delta}_m
\sim P(\mathbf{y}_m \mid \boldsymbol{\eta}_m, \boldsymbol{\delta}_m
) \label{al1}\\
& \text{ \emph{Individual latent field}: } &
\{\boldsymbol{\eta}_m, \boldsymbol{\delta}_m\} & \mid \boldsymbol{\psi}_m \sim GP(\boldsymbol \mu_m, K_m \mid \boldsymbol{\zeta}_m)\label{al2}\\
& \text{ \emph{Individual parameter priors}: } &
\boldsymbol{\zeta}_m & \mid \boldsymbol{\psi} \sim P(\boldsymbol{\zeta}_m \mid \boldsymbol{\psi})\label{al3}\\
& \text{ \emph{Global parameters priors}: } &
\boldsymbol{\psi}& \sim P(\boldsymbol{\psi})\label{al4}
\end{align}
We start by describing the model component for each individual $m$, demonstrated for $m=1$ and $m=M$ in Figure \ref{fig:DAG}. Given the individual latent field $\{\boldsymbol{\eta}_m, \boldsymbol{\delta}_m\}$, the observations are  assumed independent. There are possible parameters in the likelihood function, and we will use independent Gaussian noise $N(0,\sigma^2)$. 
The parameter $\sigma^2$ is suppressed from the notation in this section due to readability. 
The latent field consists of the physical model $\boldsymbol{\eta}_m$ and the discrepancy $\boldsymbol{\delta_m}$. Examples of these are given in Figure \ref{fig:DAG}. We define $\boldsymbol{g}_m=\{\boldsymbol{\eta}_m,\boldsymbol{\delta}_m\}$. 
If we have a model formulated as in Equation \eqref{eq:higdon}, we have
$\mathbf{g}_m \sim GP(\eta(\mathbf{X}_m, \boldsymbol{\phi}_m), K_\delta(\mathbf{X}_m,\mathbf{X}_m\mid \boldsymbol{\omega}_m))$ with $K_\delta$ and parameters as described in Section 2.1. 
If the PI prior formulation in Equation \eqref{eq:PI} is used, we have
$\mathbf{g}_m \sim 
GP(\boldsymbol{\mu}(\mathbf{X}_m\mid \boldsymbol{\beta}_m, \boldsymbol{\phi}_m), 
\mathbf{K}_{\text{disc}}(\mathbf{X}_m, \mathbf{X}_m\mid \boldsymbol{\theta}_m, \boldsymbol{\phi}_m,\boldsymbol{\omega}_{m})),$ 
with $\mathbf{K}_{\text{disc}}$ and parameters as described in Section 2.2.
The vector of all individual parameters for individual $m$ is denoted $\boldsymbol{\zeta}_m$. If the model is formulated as in Equation \eqref{eq:higdon} the vector $\boldsymbol{\zeta}_m$ consists of the physical parameters $\boldsymbol{\phi}_m$ and the discrepancy kernel hyperparameters $\boldsymbol{\omega_{m}},$ $\boldsymbol{\zeta}_m=(\boldsymbol{\phi}_m, \boldsymbol{\omega_{m})}$. 
When the physics informed priors formulation in Equations \eqref{eq:PI} is used the vector $\boldsymbol{\zeta}_m$ in addition includes the kernel and mean parameters $(\boldsymbol{\theta}_m,\boldsymbol{\beta}_m),$ and consequently $\boldsymbol{\zeta}_m = (\boldsymbol{\phi}_m, \boldsymbol{\theta}_m,\boldsymbol{\beta}_m, \boldsymbol{\omega}_{m}).$ 

The individual parameters are assumed to be conditionally independent given the global parameters $\boldsymbol{\psi}$. Further, the global parameter is given a prior distribution.
Hence, all these $M$ individual models are connected through the global parameters. 

Due to the conditional independence in the hierarchical model, the joint density of the model decomposes as follows
\begin{multline*} 
  P(\mathbf{y}_{1}, \ldots, \mathbf{y}_{M}, \mathbf{g}_1,\ldots, \mathbf{g}_M, \boldsymbol{\zeta}_{1}, \ldots, \boldsymbol{\zeta}_{M}, \boldsymbol{\psi}) = \\ 
    P(\boldsymbol{\psi}) \cdot 
    \prod_{m = 1}^{M} P(\mathbf{y}_{m} \mid \mathbf{g}_m)  \cdot 
    P(\mathbf{g}_m \mid \boldsymbol{\zeta}_m)  \cdot 
    P(\boldsymbol{\zeta}_m \mid \boldsymbol{\psi}),
\end{multline*}
which means that the individual models  are conditionally independent given the global parameters $\boldsymbol{\psi}$.

\subsection{Hierarchical Physical Model, Common Discrepancy and Shared Global Parameters}
In some cases, the discrepancy can be (almost) similar for all individuals, and hence we want to assume common discrepancy parameters for all individuals. The $m$ discrepancies are modelled as  Gaussian processes with identical  hyperparameters.
This corresponds to $\delta_m(\mathbf{X}_m)\sim GP(0, K(\mathbf{X}_m,\mathbf{X}_m \mid \boldsymbol{\omega})),$ with discrepancy hyperparameters $\boldsymbol{\omega}_{m} = \boldsymbol{\omega}$ that are identical for all the individuals. Note that the discrepancies are not assumed to take (almost) the same values, but to be realizations from the same GP, e.g. the discrepancies have the same range and marginal variance.

\subsection{Inference}
Assume we have observations from $M$ individuals where, $\mathbf{y}_{\text{pop}}= (\mathbf{y}_1,\ldots, \mathbf{y}_M)$ and the corresponding inputs are $\mathbf{X}_{\text{pop}} = (\mathbf{X}_1,\ldots, \mathbf{X}_M).$ The posterior distribution of the unknown parameters $\boldsymbol{\zeta} = (\boldsymbol{\zeta}_1,\ldots,\boldsymbol{\zeta}_M),$ $\mathbf{g}=(\mathbf{g}_1,\ldots, \mathbf{g}_m),$ $\boldsymbol{\psi}$ and $\sigma$ is given by the following equation 
\begin{multline} \label{eq:post}
  P(\mathbf{g}, \boldsymbol{\zeta}, \boldsymbol{\psi}\mid \mathbf{y}_{\text{pop}}, \mathbf{X}_{\text{pop}})  \propto \\
    P(\boldsymbol{\psi}) \cdot \prod_{m = 1}^{M} P(\mathbf{y}_{m} \mid \mathbf{g}_m, \mathbf{X}_m )\cdot 
    P(\mathbf{g}_m \mid  \boldsymbol{\zeta}_m, \mathbf{X}_m)  \cdot 
    P(\boldsymbol{\zeta}_m \mid \boldsymbol{\psi}).
\end{multline}
This posterior distribution is analytically intractable, and we rely on sampling methods. 
Since the dimension of the posterior is considerably large, traditional MCMC methods, for example, a Metropolis within Gibbs implementation, can have slow mixing and fail to converge in practice. Hamiltonian Monte Carlo \citep{neal2011mcmc, betancourt2017conceptual} provides an efficient alternative for sampling high dimensional spaces (see for example \citep{piironen2017sparsity}). The complex funnel-shape geometry of the posterior of hierarchical models can be hard to sample, while non-center parametrization can alleviate this problem  \citep{betancourt2015hamiltonian}, and it is used in this paper. More specifically, we use the NUTS \citep{hoffman2014no} variation of HMC, implemented in the probabilistic language STAN \citep{carpenter2017stan}. 
More information about the prior specification of the models and sampling can be found in the Supplementary material.

If we assume for simplicity that each of the $M$ individuals has $n$ observations, the computational complexity of the model is $\mathcal{O}(M \cdot n^3).$ Where the cubic complexity of the Gaussian process is scalable since $n$ is typically relatively small, and the complexity increases just linearly with the number of individuals $M.$

\section{Toy Example} \label{sec:toy}
We consider a conceptually simple model with one input parameter $x$ and one physical parameter $u$ which has been used in the literature \citep{bayarri2007framework}. The model represents the reality $\mathcal{R}$ and the misspecified model, $\eta$ are the following exponential models
\begin{align*}
    y^R(x) &= 3.5\cdot \exp(-u\cdot x)+b+\varepsilon\\
    \eta(x,u) &= 5\cdot \exp(-u\cdot x),
\end{align*}
where $\varepsilon$ is a Gaussian noise term. For noise-free data and for a given value of the physical parameter, $u=u_0,$, the discrepancy between the two models is $-1.5\cdot \exp(-u_0\cdot x)+b.$ However, we do not know the functional form in practice and thus assume a zero mean Gaussian process prior to describe the discrepancy function, $\delta(x)\sim GP(0,K_{\delta}(x,x'\mid \boldsymbol{\omega})).$ 
\begin{figure}[h!]
        \centering
	\includegraphics[scale=0.55]{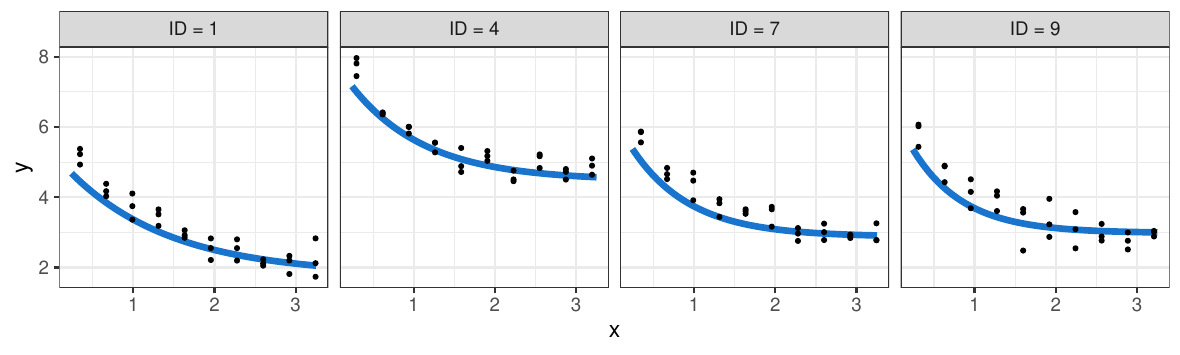} 
	\caption{Toy example: Simulated noisy data for four of the ten individuals in the data set. The blue line is the true model ($3.5\cdot \exp(-u_m\cdot x_m)+b_m$), and the dots are the corresponding noisy observed data $y^{\mathcal{R}}$.}  \label{fig:toy_data}
\end{figure}

For the simulation study, we assume there are $M=10$ individuals. For each individual, data are simulated according to model $y^R(x)$. The true physical parameter $u_m$ is set to be $0.7+0.1\cdot m$, and hence $u_1=0.8$ and $u_{10}=1.7$. The individual offsets $b_m$ are sampled randomly from a uniform distribution on the interval $[0.5, 5].$ For all individuals $\varepsilon \sim N(0,0.3^2)$. In Figure \ref{fig:toy_data}, the true model and simulated data are plotted for four individuals.

\begin{figure}[htb!]
       \centering
        \includegraphics[scale=0.55]{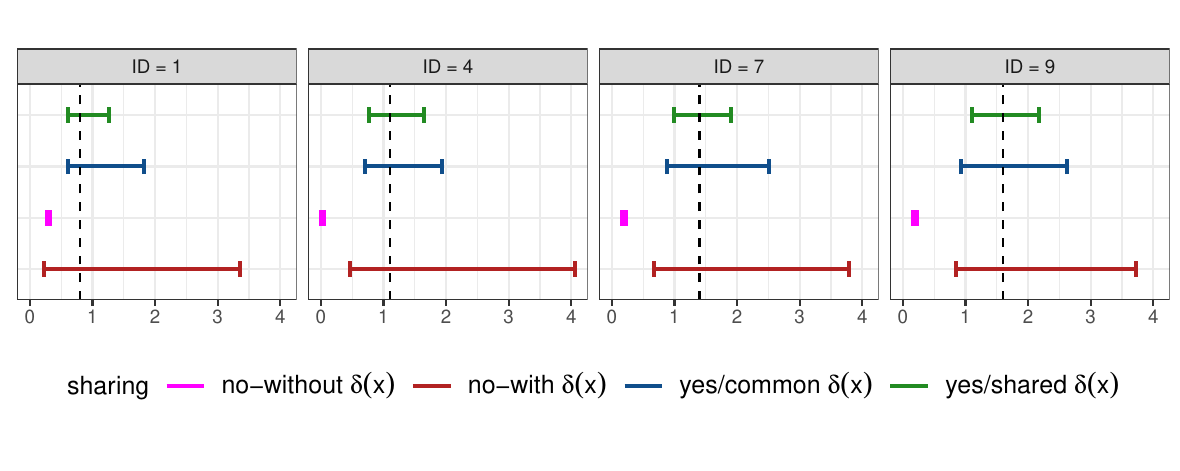} 
	\caption{Toy example: $95\%$ credible intervals for the physical parameters based on posterior distributions for the four different models. The dashed lines represent the true values of the parameter $u_m$ for each of the individuals.} 
	   \label{fig:toy_CIs}
\end{figure}

We fit four different models to the simulated data. The first model (no-without $\delta(x)$  in Figure \ref{fig:toy_CIs}), is the model $\eta$ with Gaussian noise without assuming any model discrepancy, $\delta$ and therefore is the regression model $y(x) = 5\cdot \exp(-u\cdot x) + \varepsilon,$ where $\varepsilon \sim N(0,\sigma^2).$ The second model (no-with $\delta(x)$  in Figure \ref{fig:toy_CIs}) accounts for model discrepancy and is given by equation \eqref{eq:higdon}. Both models do not share individual information and are fitted for each of the $m$ participants independently. %
The third model (yes/common $\delta(x)$  in Figure \ref{fig:toy_CIs}) shares information between individuals through a common discrepancy model and the inclusion of a global level parameter as described in Section 3.2. 
The fourth model (yes/shared $\delta(x)$ in Figure \ref{fig:toy_CIs}) shares information between the individuals through global parameters for both the discrepancy and the physical parameters as described in Section 3.1. 
It allows the discrepancies to differ between individuals, $\delta_m$ and also share information through the parameters. Furthermore, for the models that account for discrepancy, we assume that $\delta_m(x_m)\sim GP(0,K_\delta(x_m,x_m')),$ and we use the squared exponential kernel $K_\delta(x_m,x_m') = \alpha_m^2 \exp\left(-\frac{(x_m-x_m')^2}{2\rho_m^2}\right).$

In Figure \ref{fig:toy_CIs}, we see the 95\% credible intervals (CIs) of the physical parameter $u_m$ for the four different model fits for four of the ten individuals ($\text{ID} = 1,4,7,9$). 
First, we observe that if we do not account for model discrepancy (no-without $\delta(x)$) the model produces biased and over-confident parameter estimates, which comes in line with \citet{brynjarsdottir2014learning}. 
The posterior credible intervals (CIs) of the independent model, which accounts for model discrepancy (no-with $\delta(x)$), cover the true values of the parameter. 
However, the uncertainties are quite large, and this can be impractical for decision purposes. The posterior CIs of the models that share information (yes/common and shared $\delta(x)$) cover the true parameter value, and they drastically reduce the posterior uncertainties. 
The model with individual discrepancies has the smallest uncertainties. This can be explained by its flexibility which allows individual discrepancies to share individual information between them without assuming common discrepancy parameters. 
In Table \ref{table:reduction}, we summarize (for all individuals) the reduction of uncertainty of the proposed method compared to the independent models that account for model discrepancy. 
The model with common discrepancy reduces the posterior uncertainty $54\%$, and the model with shared discrepancy reduces the uncertainty $70\%$ on average. 
Finally, in Table \ref{table:rmse}, we report the prediction root mean square error (RMSE) on test data using the four models. 
We see that the model without discrepancy has the highest RMSE (0.64), while the independent models that account for discrepancy have the same RMSE (0.34).
A simulation study with a larger number of individuals and more information about the models, priors, predictions and implementation can be found in the Appendix.
\begin{table}[ht] 
\centering
\resizebox{0.85\columnwidth}{!}{
\begin{tabular}{l|rrrr}
  \hline
  case study & \textcolor{magenta}{\shortstack{no \\ $\delta(\cdot)$}}& \textcolor{Firebrick4}{\shortstack{no \\with $\delta(\cdot)$}}  & \textcolor{DodgerBlue4}{\shortstack{yes \\ common $\delta(\cdot)$}} & \textcolor{ForestGreen}{\shortstack{yes \\shared $\delta(\cdot)$}} \\ 
  \hline
Toy example & 0.63 & 0.34 & 0.34 & 0.34 \\  
WK simulation & 9.49 & 2.44 & 2.38 & 2.37 \\ 
WK real data & 9.15 & 2.98 & 2.69 & 2.61 \\ 
   \hline 
\end{tabular}}
\caption{Prediction RMSE} \label{table:rmse}
\end{table}
\section{Cardiovascular Model}\label{sec:WK_synth}
In this Section, we briefly describe the Windkessel models which are low-fidelity models of the cardiovascular system. Then we consider a synthetic case where we simulate data from a more complex model than our modelling choice and we fit the four different models as in Section \ref{sec:toy}. Finally, we fit the models using real data obtained from a pilot randomized controlled trial study.

\subsection{Models}
\begin{figure}[h!]
    \centering
    \includegraphics[scale=0.5]{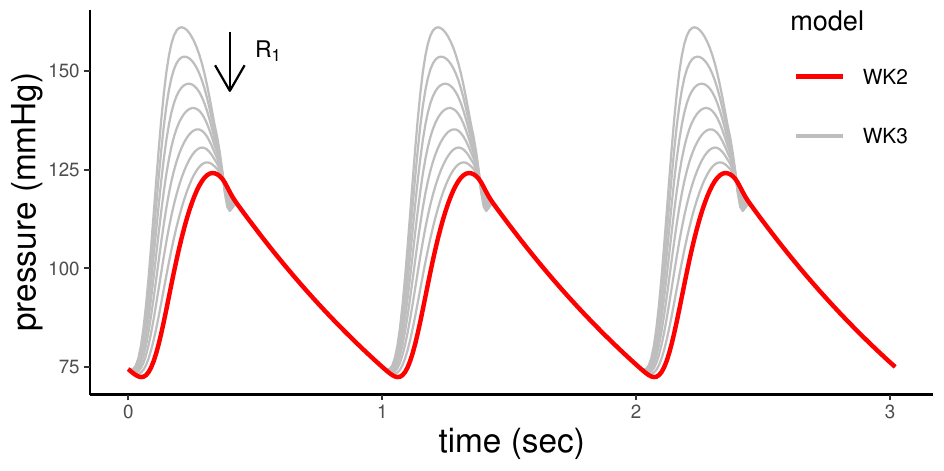} 
    \caption{Blood pressure (three cardiac cycles/heart beats) generated from the WK2 model (red) and for a range of $R_1$ values $[0.02,0.12]$ from the WK3 model (grey). The flow and $C$ values are identical for both models. The amplitude of WK3-generated curve decreases linearly with $R_1,$ while the models become equivalent for $R_1 = 0.$} \label{fig:WK23}
\end{figure}

The Windkessel (WK) models \citep{westerhof2009arterial} are linear differential equations that describe the relationship between the blood pressure, $P(t)$ and blood flow, $Q(t)$ in terms of physically interpretable parameters. The two parameters WK model (WK2) is the basis for more complex models and is given by the following time-dependent linear differential equation 
\begin{equation}
	 Q(t) = \frac{1}{R}P(t) + C \frac{dP(t)}{dt} \label{eq:WK2},
\end{equation}
where $R$ is the total vascular resistance and $C$ is the arterial compliance. These hemodynamical parameters are the physical parameters of interest.
The three parameters WK model (WK3) is described by the following differential equation, $\frac{d P(t)}{d t} + \frac{P(t)}{R_2C} = \frac{Q(t)}{C} \left (1 + \frac{ R_1}{R_2} \right ) + R_1 \frac{d Q(t)}{dt}.$ The addition of the third parameter $R_1$ can increase model flexibility and might fit the observed data better, though it overestimates the total arterial compliance $C$ \citep{segers2008three}. In practice, $R_1$ controls the amplitude of the blood pressure waveform (see Figure \ref{fig:WK23}) and therefore controls the discrepancy between the two models.
The WK2 model is a special case of the WK3 when $R_1=0$ and also we have that $R^{\text{WK2}}=R_1^{\text{WK3}}+R_2^{\text{WK3}}$ \citep{westerhof2009arterial}, and this is an important connection for the simulation study in the following section.

\subsection{Simulation Study} \label{sec:WK_sim}
\begin{figure}[h!]
        \centering
	\includegraphics[scale=0.55]{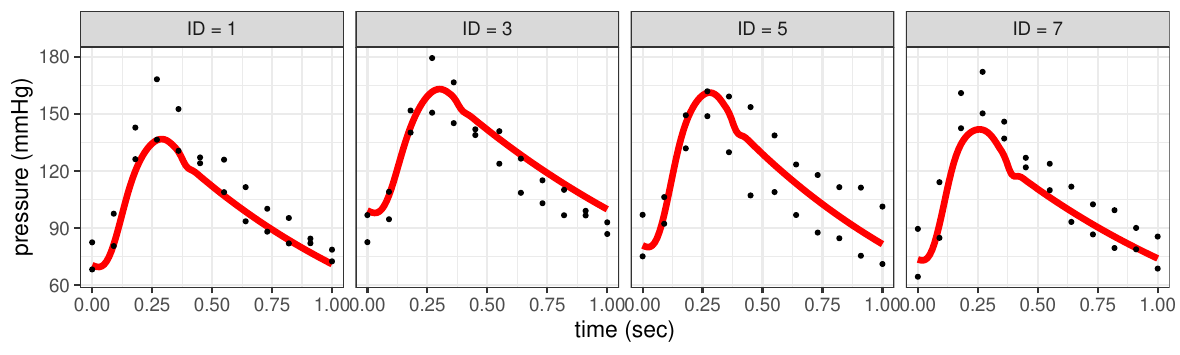} 
	\caption{Cardiovascular simulation study: Simulated noisy pressure data for four of the nine individuals in the data set. The red line is the true model (WK3), and the dots are the observed blood pressure data.}  \label{fig:wk3_data}
\end{figure}

\begin{figure}[h!]
   \centering
   \includegraphics[scale=0.55]{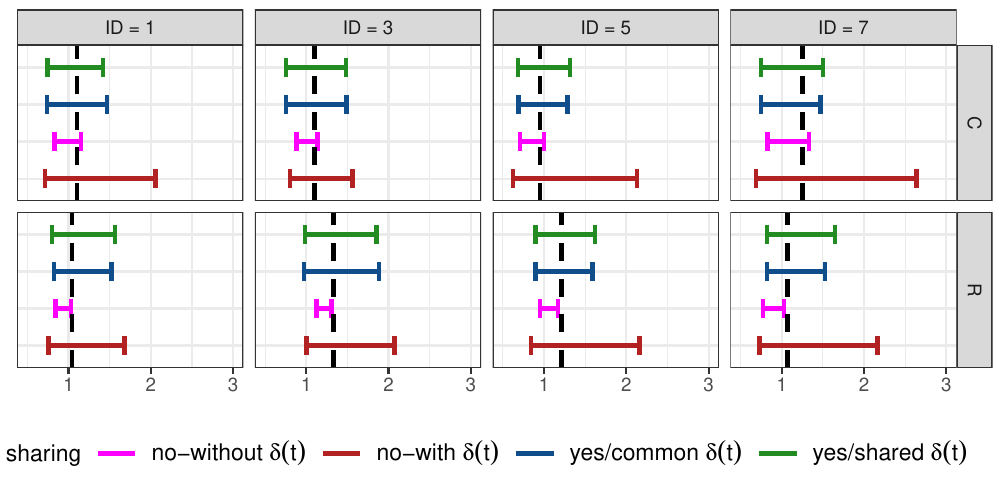} 
   \caption{Cardiovascular simulation case study: $95\%$ credible intervals for the physical parameters based on the posterior distributions for the four different models for each individual.}  \label{fig:wk_CIs}
\end{figure}

To validate our approach, we need a model that is more complex than our modelling choice but also with known connections between their physical parameters. 
Therefore, we use the WK2 model as a modelling choice ($\eta$ in Section \ref{sec:MD}) and  simulate data from the WK3 model (see Figure \ref{fig:wk3_data}), which we consider as the true model ($\mathbf{y}^\mathcal{R}$ in Section \ref{sec:MD}). We simulate data for $M=9$ individuals. For each individual $m=1,2,\ldots,9$ we use an observed flow and given individual parameters $R_{1 m},$ $R_{2 m}$ and $C_m$ to simulate pressure observations.  
The individual parameters are chosen such that we have individuals with
all the 9 possible combinations of the values $R_{2}=(1,1.15, 1.3) \text{ and } C = (0.95,1.1,1.25).$ The $R_{1m}$ parameter which controls the discrepancy between WK2 and WK3 is sampled randomly from a uniform distribution on the interval $[0.02,0.1].$
More specifically, for the blood flow $Q(t),$ we simulate the individual pressure $P_m(t) = \text{WK3}(Q(t), R_{1m}, R_{2m}, C_{m}),$ and we use $\mathbf{t}_P$ temporal locations for the pressure observations and $\mathbf{t}_Q$ temporal locations for the flow observations. The pressure observations and flow observations are not required to be aligned. 
We add i.i.d Gaussian noise to get flow and pressure observations as follows
\begin{align*}
    \mathbf{y}_{P_m} &= P_m(\mathbf{t}_P) + \varepsilon_P, \varepsilon_P \sim  N(0,4^2)\\
    \mathbf{y}_{Q_m} &= Q_m(\mathbf{t}_Q) + \varepsilon_Q, \varepsilon_Q \sim  N(0,10^2). %
\end{align*}
Our modelling choice, the WK2 model \eqref{eq:WK2} is a linear differential equation which can be written as $\mathcal{L}_{t}^{\boldsymbol{\phi}}P(t) = Q(x),$ where $\mathcal{L}_{t}^{\boldsymbol{\phi}} = \frac{1}{R} + C \frac{d}{dt}.$ By assuming that $P(t)\sim GP(\mu_P, K_{PP}(t,t'\mid \boldsymbol{\theta})),$ where $K_{PP}(t,t') = \alpha_{\text{WK2}}^2 \exp\left(-\frac{( t_P-t'_P)^2}{2\rho_{\text{WK2}}^2}\right),$ we construct a PI prior that accounts for model discrepancy as described in Section \ref{sec:PIP}, given by the formulation \eqref{eq:PI} and we fit four different models as in the Section \ref{sec:toy}. 
The first model (no-without $\delta(t)$ in Figure \ref{fig:wk_CIs}) is the PI prior without the model discrepancy, which is equivalent to model \eqref{eq:PI} but without the term $K_\delta$ in the first element of the covariance matrix. 
The second model (no-with $\delta(t)$ in Figure \ref{fig:wk_CIs}) accounts for model discrepancy and is given by equation \eqref{eq:PI}. 
Both models that do not share individual information and are fitted for each participant $m$ independently. 
The third model (yes/common $\delta(t)$ in Figure \ref{fig:wk_CIs}) shares information between individuals by assuming a common discrepancy and  including global level parameters as described in Section 3.2. 
The fourth model (yes/shared $\delta(t)$ in Figure \ref{fig:wk_CIs}) shares information between individuals through global parameters for both the model and physical parameters. 
For the models that account for discrepancy we assume that $\delta_m(t_P)\sim GP(0,K_{\delta_m}(t_P,t'_P)),$ and we use the squared exponential kernel, $K_\delta(t_P,t'_P) = \alpha^2 \exp\left(-\frac{( t_P-t'_P)^2}{2\rho^2}\right).$

In Figure \ref{fig:wk_CIs}, we see the $95\%$ credible intervals for the posterior distributions of the physical parameters $R_m$ and $C_m,$ with the corresponding true values represented by the dots. Note, that the true $R_m$ values are equal to $R_{1 m}+R_{2 m}.$ First, we observe that the model which does not account for model discrepancy (no-without $\delta(t)$), systematically underestimates both physical parameters $R$ and $C$, which comes in line with \citet{brynjarsdottir2014learning} and the results for the toy example in Section \ref{sec:toy}. 
For the $C$ parameter, the true value is within the $95\%$ CI, though it is in the upper tail of the distribution. However, in practice, the real observed data are different from the WK3 simulated data, and the fit without accounting for discrepancy might result in larger biases. The independent models that account for model discrepancy (no-with $\delta(t)$), cover the true values for all individuals, though in some cases, the uncertainty is quite large. 
For example, for individual 7, the posterior covers almost all the prior range, which is $R\sim Unif(0.5,3),$ and thus the result can have low practical value. 
The models that share information between individuals (yes/common and shared $\delta(t)$) reduce the uncertainty substantially and still cover the true parameter values. However, compared to the toy example in Section \ref{sec:toy}, the model that shares  discrepancy parameter information has similar performance to the model, which assumes the same discrepancy for all individuals. This can be understood by observing that the discrepancies between the WK3 (true model) and the WK2 (modelling choice) are very similar, thus the correlation length scale and the marginal variance ($\rho_m$ and $\alpha_m^2$ parameters in $K_\delta$), are similar for all individuals. The reduction of uncertainty of the proposed approach is summarized in Table \ref{table:reduction}, where the common discrepancy model achieves $34\%$ and $47\%$ reduction of uncertainty for $R$ and $C$, respectively. While the shared discrepancy model achieves $33\%$ and $48\%$ reduction of uncertainty for $R$ and $C$ respectively
Consequently, there is no need to consider different discrepancy parameters for each individual, and the more parsimonious parametrization of the model is sufficient. We also use the models to predict on test data, and the predictions  are summarized in Table \ref{table:rmse}. 
More details about the models, priors predictions, and implementation can be found in the supplementary material.
\begin{table}[ht] 
\centering
\resizebox{0.85\columnwidth}{!}{
\begin{tabular}{lrrr}
  \hline
  case study & parameter & \textcolor{DodgerBlue4}{yes/common $\delta(\cdot)$} & \textcolor{ForestGreen}{yes/shared $\delta(\cdot)$}\\ 
  \hline
Toy example    & u & 54 & 70\\ 
  \hline
WK simulation  & R & 34 & 33 \\ 
               & C & 47 & 48 \\ 
  \hline             
WK real data   & R & -1 & 1 \\ 
               & C & 60 & 59\\ 
   \hline
\end{tabular}}
\caption{Uncertainty reduction in \% between the individual models that account for discrepancy and the proposed method.} \label{table:reduction}
\end{table}

\subsection{Real Case Study}
\begin{figure}[h!]
   \centering
   \includegraphics[scale=0.55]{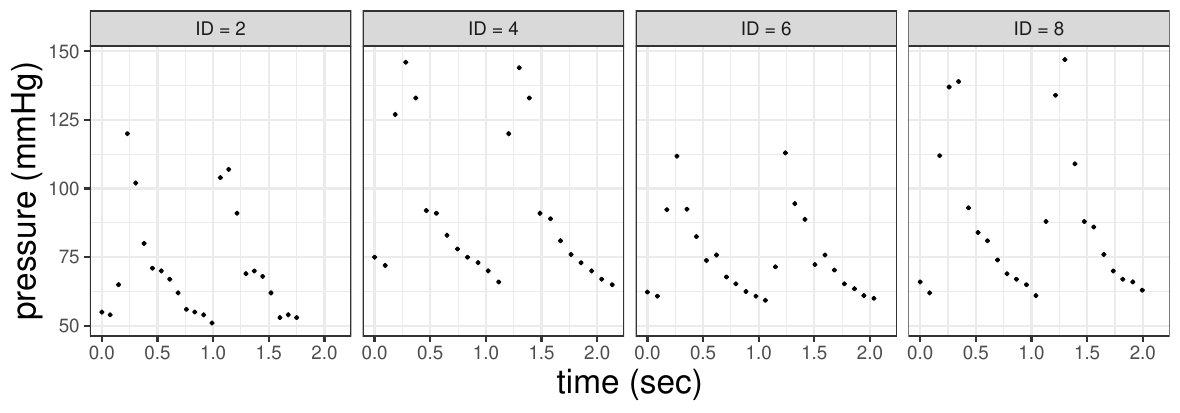} 
   \caption{Observed blood pressure data from the randomized control trial. For each participant, we have observations for two cardiac cycles, which correspond to the time duration between two consecutive heartbeats.}  \label{fig:P_trial}
\end{figure}

In this case study, we use blood pressure and blood flow data from a pilot randomized controlled trial study, which was approved and registered on clinicaltrials.org (Identifier: NCT 04151537). We have data for $M=8$ participants.
For each individual, we have 2 cycles of finger blood pressure ($n_P=24$) and aortic flow data ($n_Q=28$) (see Figure \ref{fig:P_trial}). Each cycle corresponds to the time duration between two consecutive heartbeats. 

We fit the same models as in Section \ref{sec:WK_sim}, with the main difference that we use a periodic kernel since the phenomenon repeats at each heartbeat. More specifically, we have that %
$K_{PP}(t,t') = \alpha_{\text{WK2}}^2\exp\left(-\frac{2\sin^2(\pi(t - t')/p)}{\rho_{\text{WK2}}^2}\right)$ and for the discrepancy kernel $K_{\delta}(t,t') = \alpha_{\delta}^2\exp\left(-\frac{2\sin^2(\pi(t - t')/p)}{\rho_{\delta}^2}\right).$ The heart rate is known, hence we fix the period parameter, $p.$

\begin{figure}[h!]
   \centering
   \includegraphics[scale=0.55]{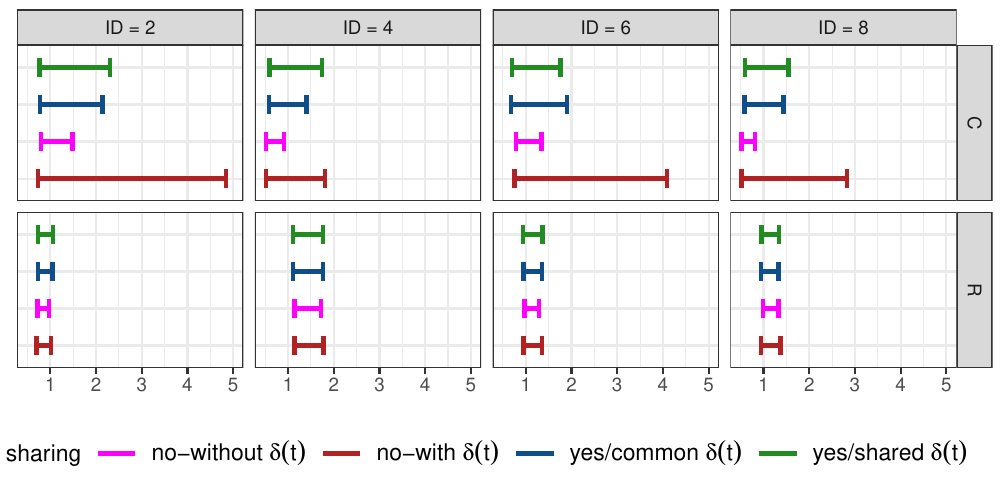} 
   \caption{Cardiovascular real data case study: $95\%$ credible intervals for the physical parameters based on the posterior distributions for the four different models for each individual. The dots represent the true individual parameter values.}  \label{fig:wk_trial_CIs}
\end{figure}

In Figure \ref{fig:wk_trial_CIs}, we see the $95\%$ CIs for the posterior distributions of the physical parameters $R_m$ and $C_m,$ for four of the eight individuals ($\text{ID}=2,4,6,8$).
For the resistance, $R$, we observe that posterior distributions for all methods are almost identical for each individual. 
In addition, the posterior uncertainty is relatively small. For the compliance, $C$, we observe that the individuals that account for model discrepancy (no-with $\delta(t)$) have quite large posterior uncertainty, where the two models of the proposed method (yes/common $\delta(t)$ and yes/shared $\delta(t)$) have reduced the uncertainty significantly. 
Similar to the simulation study, the models with common and shared discrepancy perform similarly. 

We also study the average reduction of uncertainty of the proposed approach between the models that share information and the independent models that account for model discrepancy, and the results are summarized in Table \ref{table:reduction}. For $R$, there is practically no uncertainty reduction, while for the common $\delta(t)$ there is a small increase in uncertainty ($-1\%$). However, for $C$, the uncertainty reduction for the common $\delta(t)$ model is $60\%$, and for the shared $\delta(t)$ model is $59\%.$

We use the four models to predict the following cardiac cycle on test data, and we report the average RMSE among individuals in Table \ref{table:rmse}. As in the simulation study, the three models that account for model discrepancy produce more accurate predictions compared to the model that does not account for discrepancy. While the models of the proposed method are more accurate (the RMSE are $2.69$ for the common $\delta(t)$ model and $2.61$ for the shared $\delta(t)$ model) than the independent models that account for discrepancy (RMSE of $2.98$).

\section{Discussion}
In this paper, we have developed a fully probabilistic modelling framework for learning individual physical parameters from low-fidelity models while accounting for model discrepancy and sharing knowledge between individuals. 
In the case studies, we showed that the proposed method can produce more accurate estimates of the physical parameters, reduce uncertainty significantly, and increase prediction accuracy. 
The method is scalable since the computational complexity increases linearly with the number of individuals. 
The complexity of the individual data is cubic, but the number of data for each individual is typically small. 
In cases where the individual data is large, GP methods for big data might be used (see, e.g. \cite{hensman2015mcmc, rossi2021sparse, spitieris2023bayesian}).

Digital Twin technologies have promised to transform numerous sectors of society, for example, healthcare (precision medicine), manufacturing and energy, among others \citep{DigitalTwinAdilogTrond}. 
Our method could be applied in various digital twin applications where physical models are used to gain understanding and explainability.
The most direct application of our methodology is digital twins for improving personalized healthcare \citep{corral2020digital} by using mechanistic models that encapsulate knowledge of physiology. 

\bibliography{Refs}
\newpage

\section*{Appendix}
\appendix

\section{Prior Specification}
In this Section, we specify the non-center parameterization for the general case where the model is fast to evaluate and in the case of the PI GP prior.
\subsection{General Formulation: \texorpdfstring{$y^\mathcal{R}(\mathbf{x})=\eta(\mathbf{x},\boldsymbol{\phi}) + \delta(\mathbf{x}) +\varepsilon$}{TEXT}}\label{sec:prior_general}

For simplicity, lets assume that the input $x$ and the physical parameter $\phi$ are both univariate. We take a Gaussian prior  on the physical parameter $\phi,$ $\phi \sim  N(\mu_\phi, \sigma^2_\phi)),$ and $\mu_\phi,$ $\sigma_\phi$ are the global level parameters, where $\mu_\phi\sim P(\mu_\phi)$ and $\sigma_\phi\sim P(\sigma_\phi).$ Then we use the following non-center parameterization 
\begin{equation}
    \begin{split}
        \tilde{\nu}_\phi & \sim N(0,1) \\
        \phi & = \mu_\phi + \sigma_\phi \cdot \tilde{\nu}_\phi \sim N(\mu_\phi, \sigma^2_\phi).
    \end{split}
\end{equation}
For the GP prior on the discrepancy $\delta(x) \sim GP(0,K_{\delta}(x,x'\mid \boldsymbol{\omega})),$ we use the squared-exponential kernel, $K_\delta(x,x') = \alpha^2 \exp\left(-\frac{(x-x')^2}{2\rho^2}\right)$ and $\boldsymbol{\omega }= (\alpha, \rho).$ For both parameters we use $\mathrm{Log\text{-}normal}$ priors. 
More specifically, $\rho\sim \mathrm{Log\text{-}normal}(\mu_\rho, \sigma^2_\rho),$ where  the distribution can be equivalently parameterized by the median $\textrm{m}_\rho,$   $\rho\sim \mathrm{Log\text{-}normal}(\log(\textrm{m}_\rho), \sigma^2_\rho).$  Then we use the following non-center parameterization 
\begin{equation}
    \begin{split}
        \tilde{\nu}_\rho & \sim N(0,1) \\
        \rho & = \exp(\log(\textrm{m}_\rho) + \sigma_\rho \cdot \tilde{\nu}_\rho) \sim \mathrm{Log\text{-}normal}(\log(\textrm{m}_\rho), \sigma^2_\rho)
    \end{split}
\end{equation}
and similarly for $\alpha$ we have that
\begin{equation}
    \begin{split}
        \tilde{\nu}_\alpha& \sim N(0,1) \\
        \alpha & = \exp(\log(\textrm{m}_\alpha) + \sigma_\alpha \cdot \tilde{\nu}_\alpha) \sim \mathrm{Log\text{-}normal}(\log(\textrm{m}_\alpha), \sigma^2_\alpha),
    \end{split}
\end{equation}
where $\textrm{m}_\rho\sim P(\textrm{m}_\rho), \sigma_\rho \sim P(\sigma_\rho), \textrm{m}_\alpha\sim P(\textrm{m}_\alpha), \sigma_\alpha \sim P(\sigma_\alpha).$

This can be generalized in cases where the physical parameters $\boldsymbol{\phi}$ and inputs $\mathbf{x}$ have higher dimension.
Furthermore, the global level parameters the priors are chosen in a way that reflects the population level properties.
\subsection{Physics-Informed Gaussian Process}
The main difference when we use the PI GP prior compared to \ref{sec:prior_general}, is that it involves the mean and kernel hyperparameters, $(\boldsymbol{\beta}, \boldsymbol{\theta})$ in addition to the physical parameters $\boldsymbol{\phi}$. For the mean parameters, $\boldsymbol{\beta}$ we can use the same parameterization with the physical parameter $\phi$ as in \ref{sec:prior_general}, while for the $\boldsymbol{\theta}$ parameters we can use the same parameterization we used for the discrepancy kernel in  \ref{sec:prior_general}. 
\section{Prediction Equations} \label{sec:pred_gen}
We provide prediction equations for the two modelling cases discussed in the paper. The first case is when the physical model $\eta(\mathbf{x}, \boldsymbol{\phi})$ is fast to run (e.g. a fast numerical solver). 
The second case is when we can use the physics-informed Gaussian process prior,  and therefore no numerical discretization is needed.
\subsection{General Case} \label{se:pred_eta}

Standard GP predictions formulas can be used in this case, where the mean of the GP prior is the output of the physical model $\eta(\mathbf{x}, \boldsymbol{\phi})$ (eq. (2) in Section 2.1). More specifically, for the observed inputs $\mathbf{X},$ if $\mathbf{f}\sim GP(\eta(\mathbf{X}, \boldsymbol{\phi}), K(\mathbf{X,X'})),$ at new points $\mathbf{X_*}$ the joint distribution of the noise corrupted data $\mathbf{y}=f(\mathbf{X})+\varepsilon,\varepsilon \sim N(0,\sigma^2I)$ and $f(\mathbf{X_*})=\mathbf{f_*}$ is expressed as 
\begin{equation} \label{eq:app_GP_pred}
\begin{bmatrix}\mathbf{y} \\ \mathbf{f}_*\end{bmatrix} \sim \mathcal{N}
\left(\begin{bmatrix}\eta(\mathbf{X}, \boldsymbol{\phi}) \\ \eta(\mathbf{X}_*, \boldsymbol{\phi})\end{bmatrix},
\begin{bmatrix}\mathbf{K} +\sigma^2I & \mathbf{K}_* \\ \mathbf{K}_*^T & \mathbf{K}_{**}\end{bmatrix}
\right),
\end{equation}
where $\mathbf{K} = K(\mathbf{X,X}),$ $\mathbf{K_*} = K(\mathbf{X,X_*})$ and $\mathbf{K_{**}} = K(\mathbf{X_*,X_*}).$
The conditional distribution o $p(\mathbf{f}_* \mid \mathbf{X}_*,\mathbf{X},\mathbf{y})$ is also multivariate normal and more specifically
\begin{equation}
    \begin{split}
        p(\mathbf{f}_* \mid \mathbf{X}_*,\mathbf{X},\mathbf{y}) &= \mathcal{N}(\boldsymbol{\mu}_*, \boldsymbol{\Sigma}_*) \\
        \text{where } \boldsymbol{\mu_*} &= {\eta(\mathbf{X}_*, \boldsymbol{\phi})}+ \mathbf{K}_*^T (\mathbf{K}+\sigma^2 I)^{-1} (\mathbf{y}-\eta(\mathbf{X}, \boldsymbol{\phi}))\\
        \text{and } \boldsymbol{\Sigma_*} &= \mathbf{K}_{**} - \mathbf{K}_*^T (\mathbf{K}+\sigma^2 I)^{-1} \mathbf{K}_*\,.
    \end{split}
\end{equation}
\subsection{Physics-Informed Gaussian Process Case}\label{sec:pred_pi}

We present the prediction equations for the models formulated as PI GP priors in Section 3.1 and 3.2. Suppose want to predict both functional outputs, $u$ and $f$ at new inputs $\mathbf{X}^*_u$ and $\mathbf{X}^*_f$ respectively. 
Let $\mathbf{u}^*=u(\mathbf{X}^*_u)$ to be the predictions for $u.$ The conditional distribution $p(\mathbf{u}^* \mid \mathbf{X}^*_u, \mathbf{X}, \mathbf{y}, \boldsymbol{\zeta})$ is multivariate Gaussian and more specifically

\begin{equation}
    \begin{split}
        p(\mathbf{u}_* \mid \mathbf{X}_u^*, \mathbf{X}, \mathbf{y}, \boldsymbol{\zeta}_\delta) &= \mathcal{N}(\boldsymbol{\mu}_u^*, \boldsymbol{\Sigma}_u^*) \\
        \boldsymbol{\mu_u^*} &= \mu_u\mathbf{(X_u^*)}+ \mathbf{V}_u^*{^T} (\mathbf{K_{\text{disc}}}+\mathbf{S})^{-1} (\mathbf{y}-\boldsymbol{\mu})\\
        \boldsymbol{\Sigma_u^*} &= K_{uu}(\mathbf{X}_u^*,\mathbf{X}_u^*)+K_{\delta}(\mathbf{X}_u^*,\mathbf{X}_u^*) - \mathbf{V}_u^*{^T} (\mathbf{K_{\text{disc}}}+\mathbf{S})^{-1} \mathbf{V}_u^*,
    \end{split}
\end{equation}
where $\mathbf{V}_u^*{^T} = \begin{bmatrix}  K_{uu}(\mathbf{X}_u^*,\mathbf{X}_u)+K_{\delta}(\mathbf{X}_u^*,\mathbf{X}_u) &  K_{uf}(\mathbf{X}_u^*,\mathbf{X}_f) \end{bmatrix}.$ 
The conditional distribution $p(\mathbf{f}_* \mid \mathbf{X}_f^*, \mathbf{X}, \mathbf{y}, \boldsymbol{\zeta})$ is multivariate Gaussian and more specifically
\begin{equation}
    \begin{split}
        p(\mathbf{f}_* \mid \mathbf{X}_f^*, \mathbf{X}, \mathbf{y}, \boldsymbol{\zeta}_{\delta}) &= \mathcal{N}(\boldsymbol{\mu}_f^*, \boldsymbol{\Sigma}_f^*) \\
        \boldsymbol{\mu_f^*} &= \mu_f\mathbf{(X_f^*)}+ \mathbf{V}_f^*{^T} (\mathbf{K_{\text{disc}}}+\mathbf{S})^{-1} (\mathbf{y}-\boldsymbol{\mu})\\
        \boldsymbol{\Sigma_f^*} &= K_{ff}(\mathbf{X}_f^*,\mathbf{X}_f^*) - \mathbf{V}_f^*{^T} (\mathbf{K_{\text{disc}}}+\mathbf{S})^{-1} \mathbf{V}_f^*,
    \end{split}
\end{equation}
where $\mathbf{V}_f^*{^T} = \begin{bmatrix}  K_{fu}(\mathbf{X}_f^*,\mathbf{X}_u) &  K_{ff}(\mathbf{X}_f^*,\mathbf{X}_f) \end{bmatrix}.$
\section{Hierarchical Physical Model, Common Discrepancy and Shared Global Parameters (Details on Section 3.2)}
This is the case that inputs are observed at the same domain and discrepancy is expected to have similar characteristics. 
For example, in 1D  and for the squared exponential kernel, $K_\delta(x_m,x_m') = \alpha_m^2 \exp\left(-\frac{(x_m-x_m')^2}{2\rho_m^2}\right)$ it means that the correlation decay characterized by $\rho_m$ and the marginal variance $\alpha_m^2$ is similar for all individuals, thus $\rho_m=\rho$ and $\alpha_m^2=\alpha^2$ for $m=1,\ldots, M.$ 
Therefore, $\rho$ and $\alpha$ are not controlled by global parameters and their priors distributions have fixed parameter values. 

For simplicity, lets consider the case that the physical model $\eta$ is fast to evaluate and the input and the physical parameters are univariate and we have the following formulation $y^\mathcal{R}(x)=\eta(x,\phi)+\delta(x)+\varepsilon$.  
We follow the notation of Section 3.1 and for individual input data $\mathbf{X}_m$ we have the latent field $\mathbf{g}_m \sim GP(\eta(\mathbf{X}_m, \boldsymbol{\phi}_m), K_\delta(\mathbf{X}_m,\mathbf{X}_m\mid \boldsymbol{\omega}_m)).$ In this case the vector of individual parameters $\boldsymbol{\zeta}_m$ consist of the individual physical parameter $\phi_m$ and the kernel hyperparameters $\boldsymbol{\omega}_m,$  $\boldsymbol{\zeta}_m = (\phi_m, \boldsymbol{\omega}_m).$ In Section 3.1 both have priors $\phi_m \sim P(\phi_m \mid a_\phi, b_\phi), \boldsymbol{\omega}_m \sim P(\boldsymbol{\omega}_m \mid \mathbf{a}_{\boldsymbol{\omega}}, \mathbf{b}_{\boldsymbol{\omega}})$ that depend on the global parameters $a_\phi, b_\phi,\mathbf{a}_{\boldsymbol{\omega}}, \mathbf{b}_{\boldsymbol{\omega}},$ where  $a_\phi\sim P(a_\phi), b_\phi\sim P(b_\phi), \mathbf{a}_{\boldsymbol{\omega}}\sim P(\mathbf{a}_{\boldsymbol{\omega}}) \text{ and } \mathbf{b}_{\boldsymbol{\omega}}\sim P(\mathbf{b}_{\boldsymbol{\omega}}).$ 
Now if we assume the same discrepancy across the individuals $m=1,\ldots,M,$ we have again that $\phi_m \sim P(\phi_m \mid a_\phi, b_\phi)$, but  $\mathbf{b}\sim P(\mathbf{b}).$ 
More specifically if we use the squared exponential kernel for the discrepancy GP model, we have that $\boldsymbol{\omega} = (\rho, \alpha)$ and the joint density from Section 3.1 decomposes as follows
\begin{equation*}
    P(a_\phi) P(b_\phi) \cdot 
    \prod_{m = 1}^{M} P(\mathbf{y}_{m} \mid \mathbf{g}_m)  \cdot
    P(\mathbf{g}_m \mid \phi_m, \rho, \alpha)  \cdot
    P(\phi_m \mid a_\phi, b_\phi),
\end{equation*}
where $\rho \sim P(\rho), \alpha\sim P(\alpha).$
\section{Toy Example (Section 4)}
In this Section, additional results for the toy example in Section 4 are presented. 

\subsection{Posteriors for Individual Parameters}
\begin{figure}[h!]
    \centering
    \includegraphics[scale=0.65]{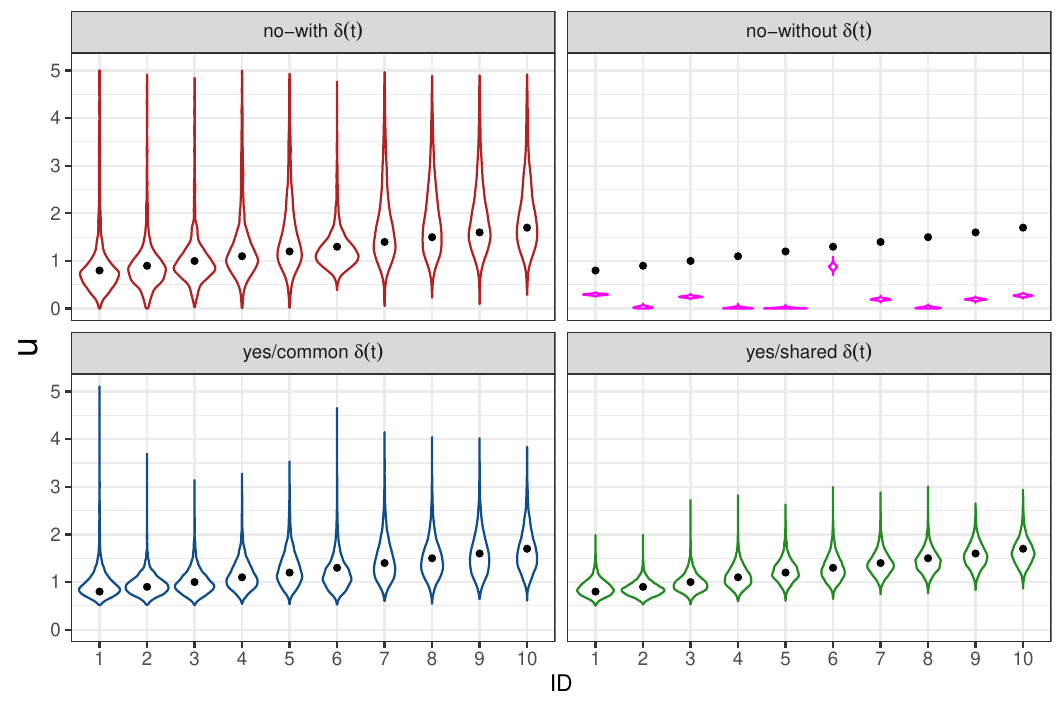} 
    \caption{Toy example, posteriors.}  \label{fig:post_toy}
\end{figure}

In Figure \ref{fig:post_toy}, we plot the posterior distributions for the physical parameter of interest u for all $M$ individuals (ID) and the four different approaches. First, we observe that if we do not account for model discrepancy (no-without $\delta(x)$), the posterior distributions for all individuals are biased and do not cover the true value. The other three models which account for model discrepancy cover the true value; however, the model that does not share information (no-with $\delta(x)$) has too large uncertainty. In contrast, the other two models share information on the physical parameter with common discrepancy (yes/common $\delta(x)$) or shared information on the discrepancy (yes/ shared $\delta(x)$), reducing the posterior uncertainty. The latter is more flexible, allows for different discrepancies, and has the smallest uncertainty.

\subsection{Predictions}
\begin{figure}[h!]
    \centering
    \includegraphics[scale=0.6]{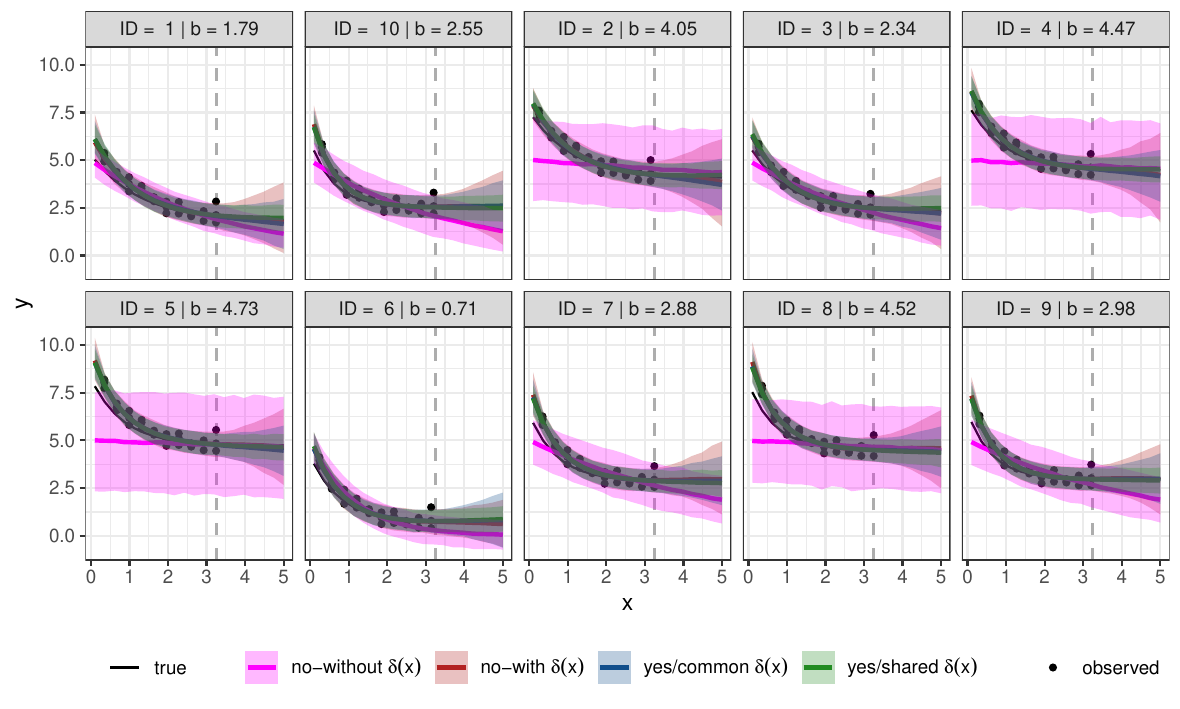} 
    \caption{Toy example, predictions.}  \label{fig:pred_toy}
\end{figure}

We now present the predictions obtained for all individuals as described in \ref{sec:pred_gen}.
In Figure \ref{fig:pred_toy}, we plot the predictions for all four approaches. We have two regions of predictions, on the left of the vertical dashed line where we have observed data (interpolation) and on the right where we have not observed data (extrapolation). First, we observe that the model which does not account for model discrepancy (no-without $\delta(x)$) does not fit the observed data well, and the prediction uncertainty is quite large. The other three approaches which account for model discrepancy perform similarly in the regions where we have observed data. While all three models perform well in regions without observed data, the model that shares information about both the physical parameter $u$ and the discrepancy $\delta,$ has the smallest uncertainty.
\section{Cardiovascular Model (Section 5)}
\subsection{Physics-Informed GP Model}
We provide details about the construction of the physics-informed prior WK model.
The model is formulated as in Section 2.2, eq.(3) as follows
\begin{equation} 
	\begin{split}
	  y_P & = P^{\text{WK2}}(t_P) + \delta(t_P) + \varepsilon_P\\
	  y_Q & = Q^{\text{WK2}}(t_Q) + \varepsilon_Q, 
	\end{split}
\end{equation}
where $P^{\text{WK2}}(t_P) \sim GP(\mu_P, K(t_P, t_P')), \varepsilon_P \sim N(0, \sigma_P^2)$ and $\varepsilon_Q \sim N(0,\sigma_Q^2)$. In addition we assume a GP prior for the model discrepancy $\delta(t_P),$ $\delta(t_P) \sim GP(\mu_P, K_{\delta}(t_P, t_P')),$ resulting in the following multi-output GP prior
\begin{equation}
p(\mathbf{y}\mid \boldsymbol \theta,  \boldsymbol \phi, \sigma_P, \sigma_Q) = \mathcal{N}(\boldsymbol{\mu}, \mathbf{K}) 
\end{equation}
where 

\begin{equation} 
    \begin{split}
        \bf{y} &= \begin{bmatrix}  \bf{y}_P \\  \bf{y}_Q \end{bmatrix},\\
        \boldsymbol{\mu} &= \begin{bmatrix}  \boldsymbol{\mu}_P \\   R^{-1} \cdot \boldsymbol{\mu}_P \end{bmatrix} \\
        \mathbf{K} &=
        \begin{bmatrix}K_{PP}(\mathbf{t}_P, \mathbf{t}_P \mid \boldsymbol \theta)+K_{\delta}(\mathbf{t}_P,\mathbf{t}_P\mid \boldsymbol \omega) + \sigma_P^2 I_P & K_{PQ}(\mathbf{t}_P, \mathbf{t}_Q \mid \boldsymbol \theta,  \boldsymbol \phi)\\
        K_{QP}(\mathbf{t}_Q, \mathbf{t}_P \mid \boldsymbol \theta,  \boldsymbol \phi) & K_{QQ}(t\mathbf{t}_Q, \mathbf{t}_Q \mid \boldsymbol \theta,  \boldsymbol \phi) + \sigma_Q^2 I_Q
        \end{bmatrix}
    \end{split}
\end{equation}
and 
\begin{equation} 
    \begin{split}
        K_{PQ}(t,t') &=  R^{-1} K_{PP}(t,t')+ C \frac{\partial K_{PP}(t,t')}{\partial t'}\\
        K_{QP}(t,t') &=  R^{-1} K_{PP}(t,t')+ C \frac{\partial K_{PP}(t,t')}{\partial t}\\
        K_{QQ}(t,t') &=  R^{-2} K_{PP}(t,t')+ C^2 \frac{\partial^2 K_{PP}(t,t')}{\partial t \partial t'}.
    \end{split}
\end{equation}
 The PI GP model which does not account for model discrepancy is the same with the difference that from the first element of the covariance matrix $\mathbf{K},$ we remove the discrepancy kernel $K_\delta.$
 
\subsection{Simulation Study}
\begin{figure}[h!]
    \centering
	\includegraphics[scale=0.6]{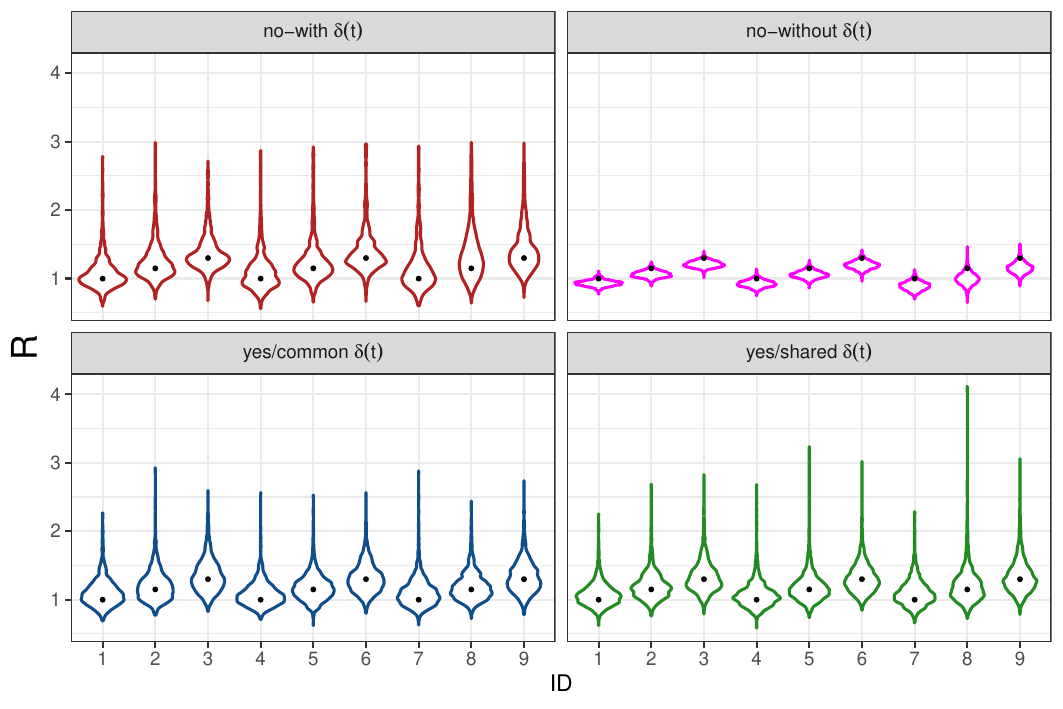} 
	\caption{Cardiovascular model: Simulation study, posterior distribution, R.}  \label{fig:post_R}
\end{figure}

\begin{figure}[h!]
    \centering
	\includegraphics[scale=0.6]{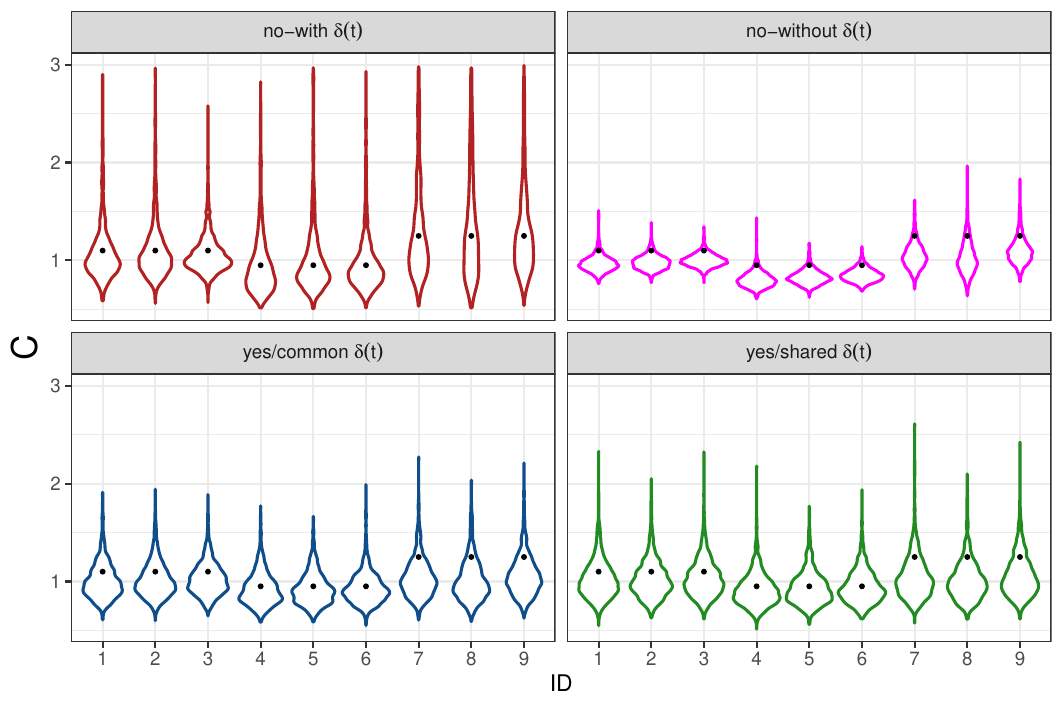} 
	\caption{Cardiovascular model: Simulation study, posterior distribution,  C.}  \label{fig:post_C}
\end{figure}

In Figures \ref{fig:post_R} and \ref{fig:post_C}, we plot the posterior distributions for $R$ and $C$ for  all four approaches. In both Figures, we observe that for the model without discrepancy (no-without $\delta(t)$), the posteriors are over-confident and underestimate both $R$ and $C.$ The model which accounts for model discrepancy but does not share information (no-with $\delta(t)$) produces more reasonable estimates of the physical parameters, though in some cases, it can be too uncertain. For example, in Figure \ref{fig:post_C} the posterior can cover the whole range of possible values. The models that share information (yes/common $\delta(t)$ and yes/shared $\delta(t)$) have reduced posterior uncertainty while covering the true values of $R$ and $C.$ Furthermore, posterior densities for the two models are very similar, suggesting that the more parsimonious model (yes/common discrepancy) is sufficient.    

\begin{figure}[h!]
    \centering
	\includegraphics[scale=0.6]{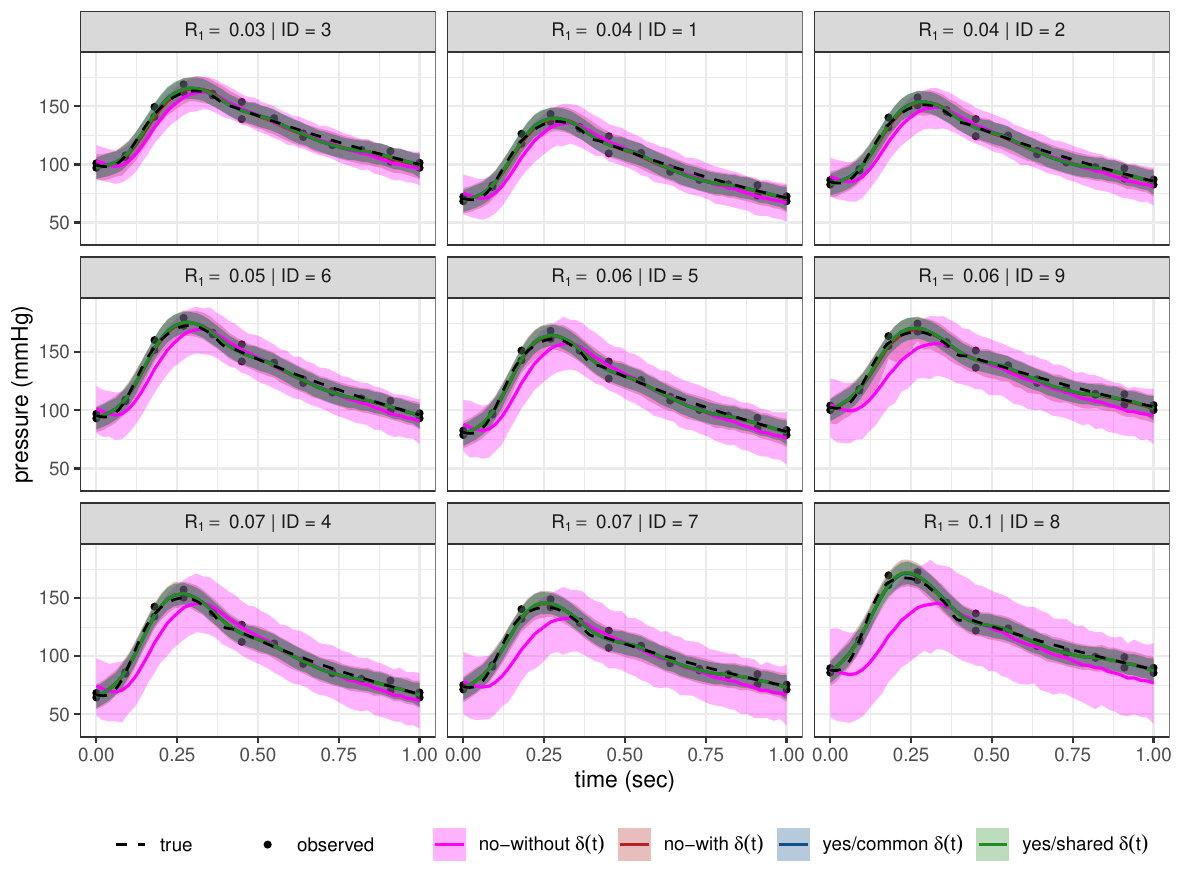} 
	\caption{Cardiovascular model, pressure predictions.}  \label{fig:pred_P}
\end{figure}

\begin{figure}[h!]
    \centering
	\includegraphics[scale=0.6]{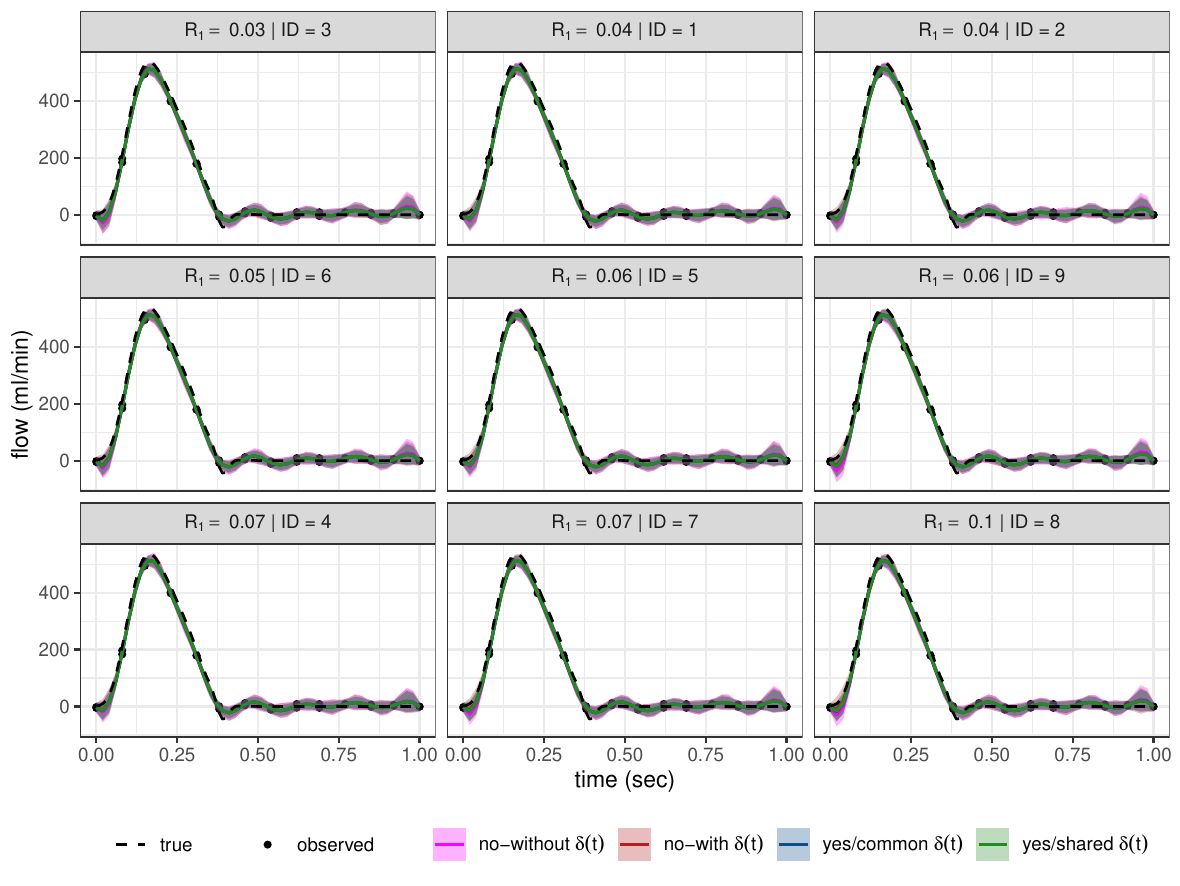} 
	\caption{Cardiovascular model, flow predictions.}  \label{fig:pred_I}
\end{figure}

In Figures \ref{fig:pred_P} and \ref{fig:pred_I}, the predictions for all four models for the two model outputs (blood pressure and blood flow) are plotted. The model which does not account for discrepancy produces more uncertain predictions than the other three models. We see that this uncertainty increases with the values of the parameter $R_1,$ which controls the discrepancy between the two models (WK2 and WK3). 
Even if the posterior uncertainty for $R$ and $C$ is larger for the model that does not share information (no-with $\delta(t)$), its prediction uncertainty is similar to the two models that share information (yes/common $\delta(t)$ and yes/shared $\delta(t)$). 
In Figure \ref{fig:pred_I}, the blood flow predictions are plotted. All models perform similarly, but the model that does not account for discrepancy (no-without $\delta(t)$) has slightly larger prediction uncertainty. 

\subsection{Real Case Study}
\begin{figure}[h!]
    \centering
	\includegraphics[scale=0.55]{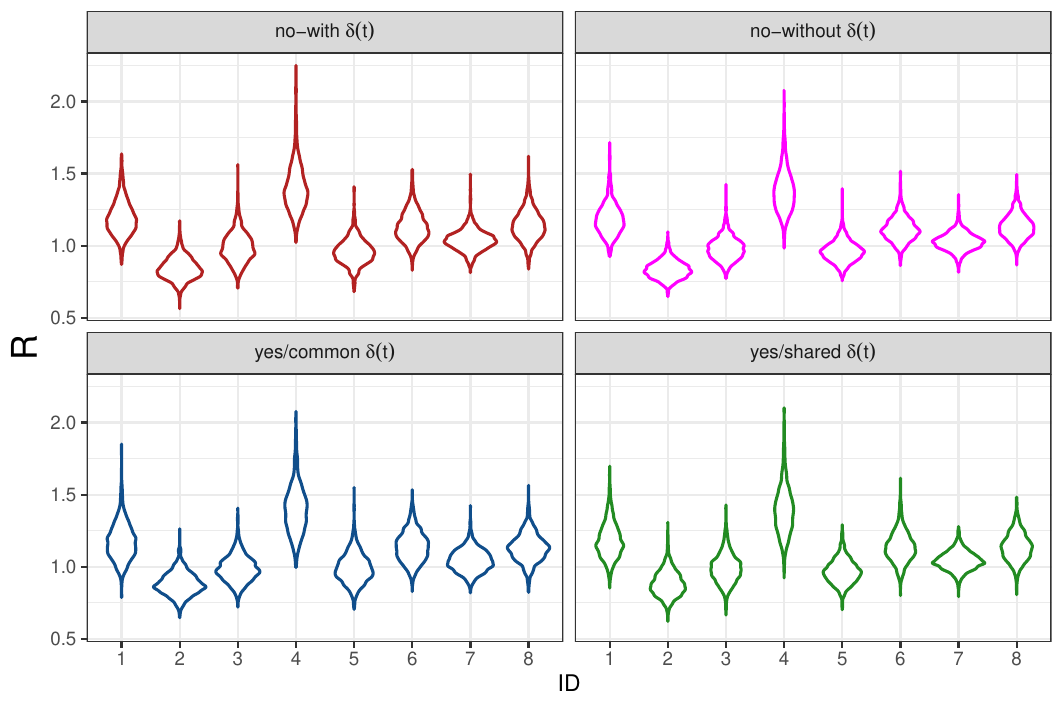} 
	\caption{Cardiovascular model: real data, posterior distribution, R.}  \label{fig:post_R_real}
\end{figure}

\begin{figure}[h!]
    \centering
	\includegraphics[scale=0.55]{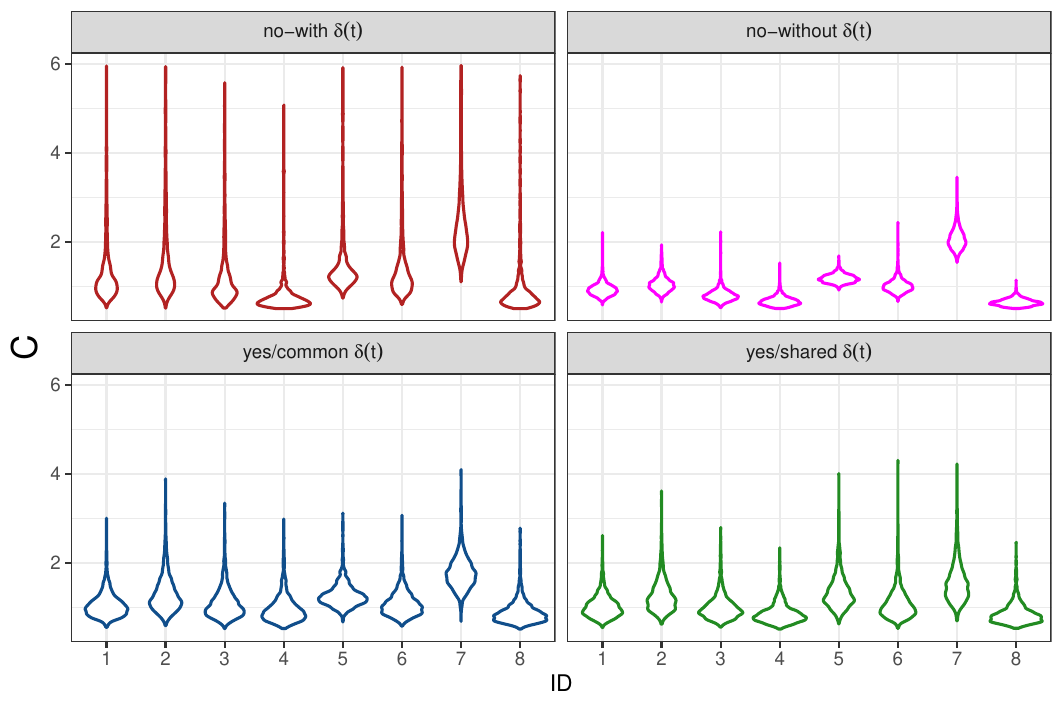} 
	\caption{Cardiovascular model: real data, posterior distribution,  C.}  \label{fig:post_C_real}
\end{figure}

In Figures \ref{fig:post_R_real} and \ref{fig:post_C_real}, the posterior distributions for $R$ and $C$ for all four approaches are plotted. 
For the physical parameter R, we observe that all four models produce quite similar posterior distributions, and the posterior variance is reasonable.
This is in contrast with the simulation study where the model no-without $\delta(t)$ was biased and overconfident. 
For parameter C, the models that share information (yes/common $\delta(t)$ and yes/shared $\delta(t)$) have reduced the posterior uncertainty significantly compared to the model that accounts for discrepancy but does not share information. This aligns with the simulation study but the reduction in uncertainty is larger in the real data. 

\begin{figure}[h!]
    \centering
 	\includegraphics[scale=0.55]{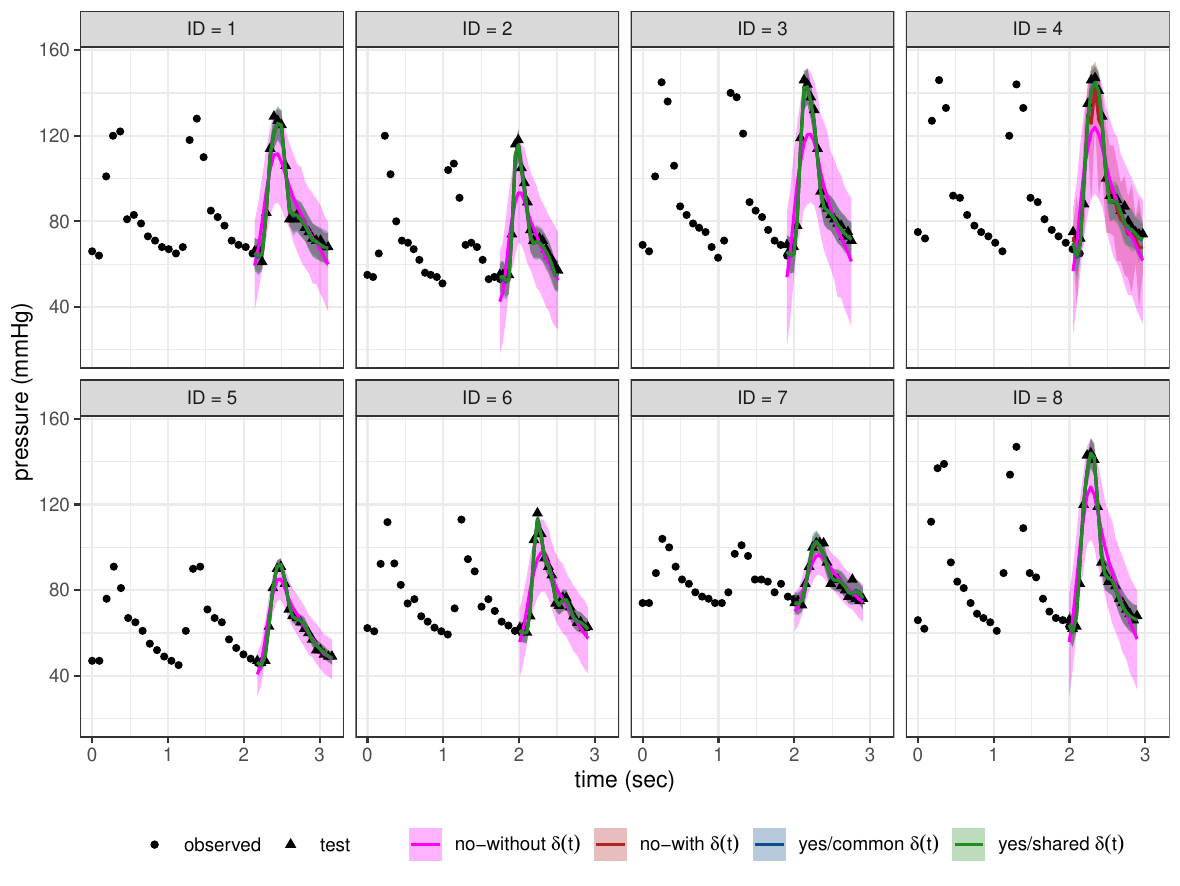} 
	\caption{Cardiovascular model: Real data, pressure predictions.}  \label{fig:pred_P_real}
\end{figure}

For each individual, there are observations from two cardiac cycles. Each cardiac cycle is the time duration between two consecutive heartbeats, and it is repeated with some physical variability. Hence, the periodic kernel is a natural choice for the physics-informed prior. Another periodic kernel is also used for the GP prior on the discrepancy. In Figure \ref{fig:pred_P_real}, the predictions for all eight individuals for all four different models are plotted. Predictions are similar to the simulation study, where we observe that the model that does not account for model discrepancy has quite large prediction uncertainty. In Table 1 in the paper, we also see the difference in the prediction RMSE, where the model that shares information between individual discrepancies is the most accurate.

\section{Larger-scale Experiment}

We now consider a larger-scale experiment by increasing the number of individuals to $M=100.$ We use the toy model again in a similar setup as described in Section 4. The true individual parameter values $u_m, m=1\ldots,100$ range again from $0.8$ to $1.7$ and the individual offsets $b_m$ are sampled randomly from a uniform distribution on the interval $[0.5, 5].$  

\begin{figure}[h!]
    \centering
	\includegraphics[scale=0.65]{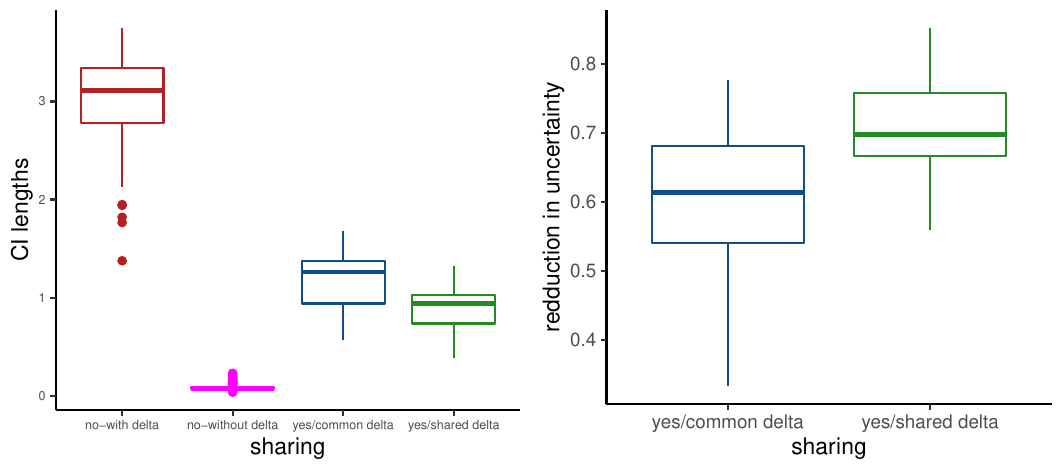} 
	\caption{Large-scale experiment with 100 individuals. Left plot presents the boxplots of $95\%$ credible interval lengths for the four models. The coverage for the model that does not account for discrepancy (no-without delta) is $0\%,$ while for the other three models that account for model discrepancy is $100\%.$ In the right plot, the reduction in posterior uncertainty is presented for the two models that share information between individuals compared to the model that accounts for discrepancy but does not share information between individuals. Both models achieve a remarkable reduction in uncertainty. The model with the common discrepancy (yes/common delta) can reduce the individual uncertainty on average $61\% (33\%,78\%),$ while the model that shares information about individual discrepancies (yes/shared delta) can reduce the individual uncertainty on average $71\%(56\%,85\%).$}  \label{fig:large_scale}
\end{figure}

As in Sections 4 and 5, we fit four models. The first model (no-without delta in Figure \ref{fig:large_scale}), is the model $\eta$ with Gaussian noise without assuming any model discrepancy, $\delta$ and therefore is the regression model $y(x) = 5\cdot \exp(-u\cdot x) + \varepsilon,$ where $\varepsilon \sim N(0,\sigma^2).$ The second model (no-with delta in Figure \ref{fig:large_scale}) 
accounts for model discrepancy and is given by equation 2 %
. Both models do not share individual information and are fitted for each of the $m$ participants independently. %
The third model (yes-common delta in Figure \ref{fig:large_scale}) shares information between individuals through a common discrepancy and  inclusion of a global level parameter as described in Section 3.2. 
The fourth model (yes-shared delta in Figure \ref{fig:large_scale}) shares information between the individuals through global parameters for both the discrepancy and the physical parameters as described in Section 3.1. 

To summarize the results for the 100 individuals, we study the coverage of the posterior distributions, that is the proportion that the posterior $95\%$ credible intervals cover the true physical parameter values. We are mainly interested in the degree of uncertainty about the physical parameter, which can be expressed by the length of the $95\%$ credible intervals, and we also study the reduction of uncertainty of the proposed approach compared to the model that accounts for model discrepancy but does not share information between individuals.

In the left plot of Figure \ref{fig:large_scale}, the boxplots of the lengths of $95\%$ credible intervals for the four models are presented. The model that does not account for model discrepancy (no-without delta) has the smallest uncertainty across all individuals, though the coverage is $0\%$. The other three models that account for model discrepancy have $100\%$ coverage. The model that does not share information between individuals has large posterior uncertainty for most individuals, and therefore even if it covers the true value, it can be quite impractical. The two models that share information between individuals (yes/common delta and yes/shared delta) have reduced the uncertainty significantly while being calibrated. The right plot of Figure \ref{fig:large_scale} presents the reductions of uncertainty for all individuals for the two models that share information between individuals compared to the model that accounts for discrepancy but does not share information. The model with the common discrepancy (yes/common delta) can reduce the individual uncertainty on average $61\% (33\%,78\%),$ while the model that shares information about individual discrepancies (yes/shared delta) can reduce the individual uncertainty on average $71\%(56\%,85\%).$

\end{document}